\documentclass[11pt,onecolumn,draftcls]{IEEEtran}

\ifCLASSOPTIONcompsoc
   \usepackage[nocompress]{cite}
\else
   \usepackage[noadjust]{cite}
\fi
\ifCLASSINFOpdf
   \usepackage[pdftex]{graphicx}
   \DeclareGraphicsExtensions{.pdf,.jpeg,.png}
\else
   \usepackage[dvips]{graphicx}
\fi
\usepackage[cmex10]{amsmath}
\usepackage{amssymb}
\usepackage{algorithmic}
\ifCLASSOPTIONcompsoc
\usepackage[tight,normalsize,sf,SF]{subfigure}
\else
\usepackage[tight,footnotesize]{subfigure}
\fi

\ifCLASSOPTIONcompsoc
  \usepackage[caption=false,font=normalsize,labelfont=sf,textfont=sf]{subfig}
\else
  \usepackage[caption=false,font=footnotesize]{subfig}
\fi
\usepackage{stfloats}
\usepackage{colortbl}
\usepackage{algorithm}
\usepackage{booktabs}

\graphicspath{{../Figs/}{}}

\numberwithin{algorithm}{section} 

\newcommand{\piZbar}{\ensuremath{\overline{\pi}_0}}
\newcommand{\piAbar}{\ensuremath{\overline{\pi}_A}}

\begin{document}

\title{Anomaly Detection in Time Series of Graphs \\ using Fusion of
  Graph Invariants}

%
%
%
%
\author{Youngser ~Park,~
        Carey~E.~Priebe,~
        and~Abdou~Youssef
\thanks{Y. Park and
  C.E. Priebe are with the Department of Applied Statistics and
  Mathematics, Johns Hopkins University, Baltimore, MD, 21211.
~See http://www.cis.jhu.edu/faculty/ for current contact information.} %
\thanks{A. Youssef is with the Department of Computer
  Science, George Washington University, Washington, D.C. 20052.}
}

\IEEEcompsoctitleabstractindextext{%

\begin{abstract}
Given a time series of graphs $G(t) = (V,E(t))$, $t=1,2,\cdots$, where
the fixed vertex set $V$ represents ``actors'' and an edge between
vertex $u$ and vertex $v$ at time $t$ ($uv \in E(t)$) represents the
existence of a communications event between actors $u$ and $v$ during
the $t^{th}$ time period, we wish to detect anomalies and/or change
points.  We consider a collection of graph features, or invariants,
and demonstrate that adaptive fusion provides superior inferential
efficacy compared to naive equal weighting for a certain class of
anomaly detection problems.  
Simulation results using a latent process
model for time series of graphs, as well as illustrative experimental
results for a time series of graphs derived from the Enron email data,
show that a fusion statistic can provide superior inference compared
to individual invariants alone. These results also demonstrate that an
adaptive weighting scheme for fusion of invariants performs better
than naive equal weighting.
\end{abstract}


}

\maketitle

\IEEEdisplaynotcompsoctitleabstractindextext

%
\IEEEpeerreviewmaketitle

\section{Introduction}
\label{sec:introduction}
%
%

%
%
%
%
\IEEEPARstart{G}{iven} a time series of graphs $G(t) = (V,E(t))$,
$t=1,2,\cdots$, 
where the vertex set $V=[n]=\{1,\cdots,n\}$ is fixed throughout
and the edge sets $E(t) \subset {{V}\choose{2}}$ are time-dependent,
we wish to detect anomalies and/or change points.
Let us consider vertices to represent ``actors,''
and an edge between vertex $u$ and vertex $v$ at time $t$ ($uv \in E(t)$)
represents the existence of a communications event between actors $u$ and $v$
during the $t^{th}$ time period.
Thus $E(t)$ represents the collection of (unordered) pairs of vertices
which communicate during $(t-1,t]$.
We will not consider
   directed edges
or hyper-graphs (hyper-edges consisting of more than two vertices)
or multi-graphs (more than one edge between any two vertices at any time $t$)
or self-loops (an edge from a vertex to itself)
or weighted edges, 
although all of these generalizations of simple graphs
may be relevant for specific applications.

\begin{figure}[!t] 
  \centering
  \includegraphics[width=3.5in]{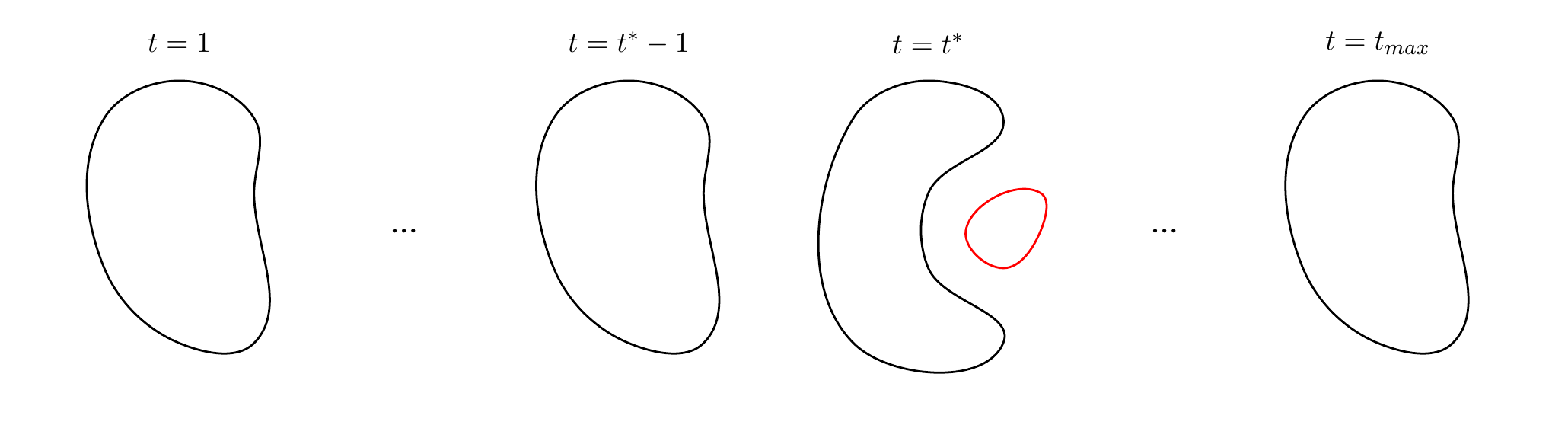}
  \caption{\label{fig:ke-ts}
  Notional depiction of a time series of graphs
  in which
  the entire vertex set $V$ behaves in some null state
  for $t=1,\cdots,t^*-1$
  and then, at time $t^*$,
  a subset of vertices $V_A$
  exhibits a change in connectivity behavior.
  }
\end{figure}

The specific anomaly we will consider is the ``chatter'' alternative --
a small (unspecified) subset of vertices with excessive communication amongst themselves
during some time period in an otherwise stationary setting,
as depicted in Figure \ref{fig:ke-ts}.
This figure notionally depicts the entire vertex set $V$ behaving in some null state
for $t=1,\cdots,t^*-1$;
then, at time $t^*$,
a collection of vertices $V_A \subset V$ ($|V_A|=m, 2 \leq m \ll n$)
exhibit probabilistically higher connectivity.
(The remaining ${{n}\choose{2}} - {{m}\choose{2}}$
interconnection probabilities remain in their null state at time $t^*$.)
Our statistical inference task is then to determine whether or not there has
emerged a ``chatter'' group at some time $t=t^*$,
as shown in Figure \ref{fig:ke-ts}.

The latent process model for time series of graphs
presented in \cite{LP2009} provides for precisely this temporal structure.
Each vertex is governed by a continuous time, finite state stochastic process
$\{X_v(t)\}_{v \in V}$, with the state-space given by $\{0,1,\cdots,K\}$.
The probability of edge $uv$ at time $t$
is determined by the inner product of the sub-probability vectors specified by
$\int_{t-1}^t I\{X_w(\tau)=k\}d\tau$, $k=1,\cdots,K$, for $w=u,v$.
For the scenario depicted in Figure \ref{fig:ke-ts},
the vertex processes
$\{X_v(t)\}_{v \in V_A}$
are stationary until time $t^*-1$ and then undergo a change point,
while the processes
$\{X_v(t)\}_{v \in V \setminus V_A}$
remain stationary throughout all time.

In \cite{LP2009}, the model produces a dependent time series of graphs $G(t)$,
each of which is itself a latent position model with conditionally independent edges
given $\{X_v(\tau)\}_{v \in V, \tau \leq t}$.
The model allows two simplifying approximations;
a second-order (central limit theorem) approximation with temporally independent random graphs
each of which is itself a random dot product (\cite{ST10,YS07}, and
Section 16.4 in \cite{BJR07})
latent position model \cite{HRH2002}, and
a first-order (law of large numbers) approximation with temporally independent random graphs
each of which is itself an independent edge random graph model
\cite{bollobas01}.

\begin{figure}[!t]
  \centering
  \includegraphics[width=2in]{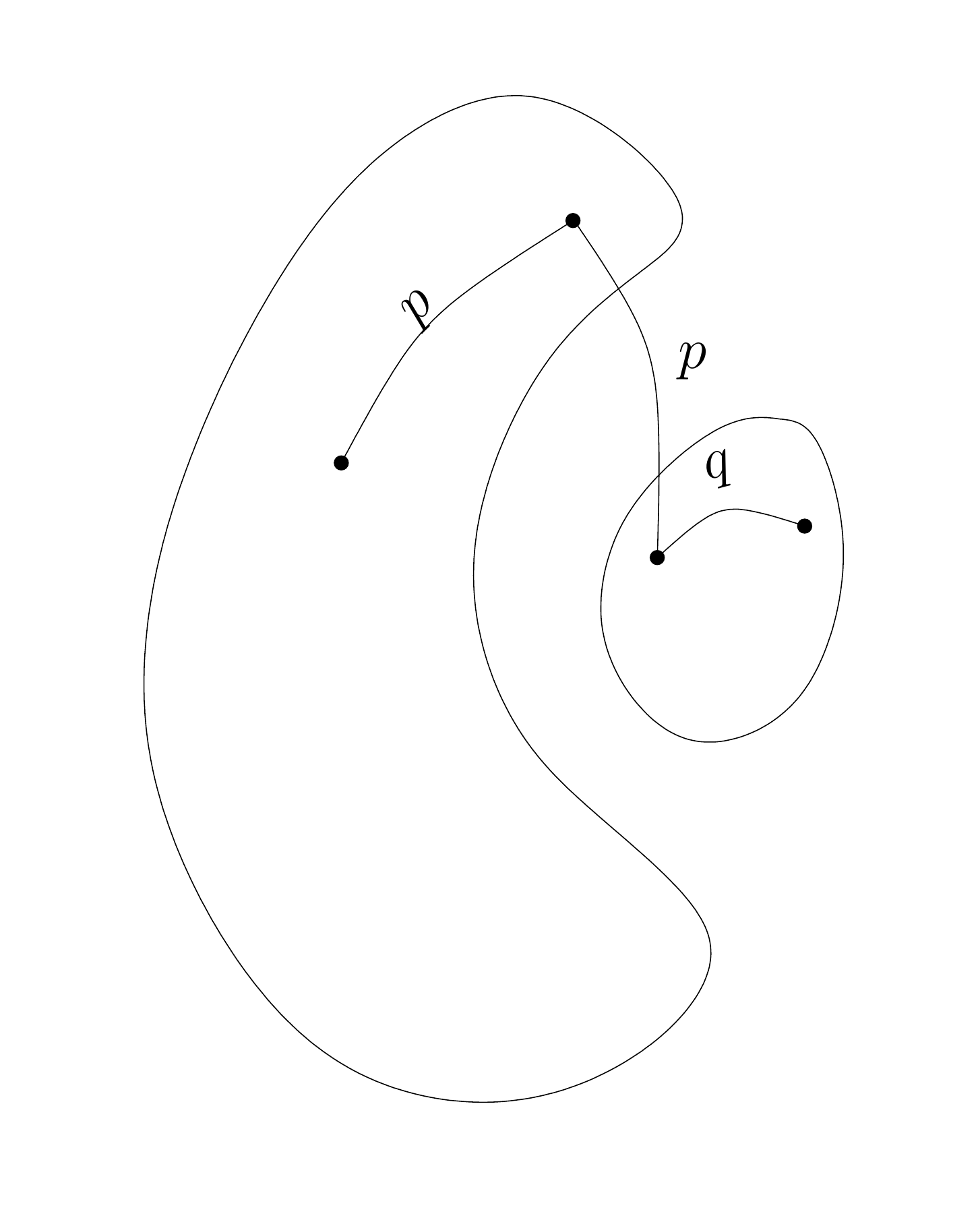}
  \caption{\label{fig:ke-one} The ``kidney-egg'' random graph
    model, denoted $\kappa(n,p,m,q)$. The small ``egg'' represents
    the $m$ vertices ($V_A$) that exhibit chatter (each edge occurring
    with probability $q$). The ``kidney'' is the population
    of $n-m$  vertices which are not exhibiting chatter (each edge
    occurring with probability $p<q$). Edges between a vertex
    in the kidney and a vertex in the egg occur with probability
    $p$. When $m=0$ or $q=p$, this model degenerates to $ER(n,p)$.
  }
\end{figure}

The simplicity of the first-order approximation, depicted in Figure \ref{fig:ke-one}
for the special case of {\em homogeneity} vs.\ {\em kidney-egg},
provides a useful framework for description.
If the vertex processes $\{X_v(t)\}_{v \in V}$ are independent and identical,
with stationary probability vector $\pi_0 = [\pi_{0,0},\pi_{0,1},\cdots,\pi_{0,K}]'$,
then the first-order approximation produces a temporally independent series of
homogeneous independent edge {\color{black}Erd\"os-R\'enyi random graph
(denoted by $ER(n,p)$)} with $p\ =\ \langle\piZbar,\piZbar\rangle$,
where $\piZbar = [\pi_{0,1},\cdots,\pi_{0,K}]'$.
The vertex processes
$\{X_v(t)\}_{v \in V_A}$
change at time $t^*-1$, taking on
stationary probability vector $\pi_A$,
so that $G(t^*)$
is a kidney-egg independent edge $\kappa(n,p,m,q)$ random graph
with $q\ =\ \langle\piAbar,\piAbar\rangle$.
The idea that the change point consists of a small collection of vertices
exhibiting {\em excessive} interconnection probability results in the
restriction of this model to the case $q>p$.
(Here we have assumed, for simplicity, that the geometry provides
$\langle\piZbar,\piAbar\rangle\ =\ p$.)\footnote{
{\color{black}
If $\langle\piZbar,\piAbar\rangle = p' \geq p$, then
we have
$\mathbb{E}[deg(v)] = m q + (n-m)\times p'$ for a $v\in egg$, and
$\mathbb{E}[deg(v)] = (n-m)\times p + m\times p'$ for a $v \in kidney$.
The difference between these two expected degrees is then
$m\times (q - p') + (n-m)\times (p' - p).$
If $m$ is of order $o(n)$, we see that the above expression is
minimized over $p' \geq p$ when $p'=p$, which indicates that
the most \textit{difficult} scenario is when $p' = p$. 
}}

In \cite{enron2005}, the scan statistic graph invariants
are introduced and applied to the problem of detecting ``chatter'' anomalies in time series of Enron graphs.
In \cite{pcp}, various graph invariants (size, maximum degree, etc.)
are considered for their power as test statistics
in testing $H_0: ER(n,p)$ vs.\ $H_A: \kappa(n,p,m,q)$.
It is demonstrated that no single invariant is uniformly most powerful.
See also \cite{newsletter}.

{\color{black}
In \cite{ide05} the principal eigenvector of a matrix based on the graph
is tracked over time, and an anomaly is declared to be present if its
direction changes by more than some threshold. Researchers in 
\cite{miller2011} have addressed problems in dynamic
network analysis such as detection of anomalies
or distinct subgraphs  in large, noisy background in signal
processing fields. 
Recently, \cite{borges2011} proposed a methodology of
detecting anomalous graphs by examining distributions of vertex
invariants instead of using a single graph invariant. They used a
simple non-time series of simulated $ER$ random graph models.
%
In \cite{neil2011}, a locality statistic using a generalized
likelihood ratio test statistic (they call this a scan
statistic) has been applied for an online network intrusion
detection. Other notable recent efforts in this direction include \cite{horn2011,sharpnack,valko}.
}

In this paper,
we consider the problem of detecting ``chatter'' anomalies in time series of graphs
using combinations of invariants.
We present experimental results for anomaly detection
on time series of simulated data from the model in \cite{LP2009},
as well as an investigation of a time series of graphs extracted from the Enron email corpus,
to demonstrate that a statistic which
combines multiple invariants can provide superior inference compared to
individual invariants alone.
We further demonstrate an adaptive weighting scheme for fusion of invariants
that performs better than naive equal weighting.

Section \ref{sec:graph-features} presents the graph features (invariants, used as statistics) considered herein,
Section \ref{sec:fusion} introduces our adaptive fusion, and
Section \ref{sec:experiments} presents results with simulated data as well as Enron email data.
We conclude with discussion in Section \ref{sec:discussion}.

\section{Graph Features}
\label{sec:graph-features}

We investigate a collection of nine graph features similar to that considered in \cite{pcp}:
size,
maximum degree,
maximum average degree (eigenvalue approximation),
scan statistic (scale 1,2,3),
number of triangles,
clustering coefficient, and
(negative) average path length.
In all cases, a
large value of the feature $F$ is an evidence in favor of
{\em excessive} interconnection probability.

\subsection{Invariants}

\subsubsection{Size}
\label{sec:size}

The size of a graph is the number of edges in the graph, given by
$$ F_1(G) = \mathtt{size}(G) = |E(G)|.$$
This is the simplest global graph statistic.

\subsubsection{Maximum Degree}
\label{sec:maximum-degree}

The maximum degree $\Delta(G)$ of a graph is given by
$$F_2(G) = \Delta(G) = \max_{v\in V}deg(v)$$
where $deg(v)$ is the degree of vertex $v$. This is the simplest localized
graph feature. 

\subsubsection{Maximum Average Degree}
\label{sec:maxim-aver-degr}

The maximum average degree of a graph is the maximum over all
subgraphs $H$ of $G$ of the average degree of $H$. 
If $deg(v)$ is the degree of vertex $v$, then the average
degree of a graph $G=(V,E)$ is 
given by
$$\bar{d}(G) = \frac{1}{|V|}\sum_{v\in V}deg(v) =
\frac{2\times \mathtt{size}(G)}{\mathtt{order}(G)}$$
where $\mathtt{order}(G)=|V|$, the number of vertices.
Thus the maximum average degree is given by
$$\mathtt{MAD}(G) = \max_{H\subset G}\bar{d}(H)$$
where the maximum is over all (induced) subgraphs $H$ of $G$.

Since $\mathtt{MAD}(G)$ is difficult to compute exactly \cite{fgt97},
we resort to an eigenvalue approximation.
$\mathtt{MAD}(G)$ is bounded above by the largest eigenvalue
of the adjacency matrix of $G$, denoted $\mathtt{MAD}_e(G)$,
and we use
$$F_3(G) = \mathtt{MAD}_e(G).$$
As demonstrated in \cite{pcp},
the eigenvalue method appears to be strictly better
at detecting increased local activity than the greedy approximation
method of \cite{fgt97} (Problem 5.7.2, page 90).

\subsubsection{Scan Statistic}
\label{sec:scan-statistic}

Scan statistics \cite{enron2005} are graph features based on local
neighborhoods of the graph. We will consider the scan statistic
$\mathtt{SS}_k(G)$ to be the maximum number of edges over all $k^{th}$ order
neighborhoods, where the $k^{th}$ order neighborhood of a vertex $v$,
$N_k[v]$,
is the set of vertices {\color{black} whose graph shortest path distance} from
$v$ is less than equal to $k$. 
We will consider
$k=\{1,2,3\}$, where $\mathtt{SS}_k(G)$ is given
by
$$F_{3+k}(G) = \mathtt{SS}_k(G) = \max_{v\in V}\mathtt{size}(\Omega(N_k[v])),$$
where $\Omega(N_k[v])$ denotes the induced subgraph.

\subsubsection{Number of Triangles}
\label{sec:number-triangles}

We consider the total number of triangles in $G$.
If $A$ is the
adjacency matrix for the graph $G$, then the number of triangles is
given by 
$$F_7(G) = \tau(G) = \frac{\mathtt{trace}(A^3)}{6}.$$
The trace is zero if and only if the graph is triangle-free.

\subsubsection{Clustering Coefficient}
\label{sec:clust-coeff}

We consider the global clustering coefficient (CC) in $G$, given by
$$F_8(G) = \mathtt{CC}(G) = \frac{ct(G)}{ot(G)},$$
where $ct$ is the number of closed triplets (a subgraph with three
vertices and three edges) and $ot$ is the number of open triplets (a
subgraph with three vertices and at least two edges). This measures
the probability that 
the adjacent vertices of a vertex are connected. This is sometimes
called the \textit{transitivity} of a graph.

\subsubsection{Average Path Length}
\label{sec:average-path-length}

The average path length (APL) is given by
$$\mathtt{APL}(G) = \frac{\sum_{u,v}s(u,v)}{n(n-1)},$$
where {\color{black}$s(u,v)$} is the shortest path between vertices $u$ and $v$.
This measures how many steps are required to access every other vertex
from a given vertex, on average. Unlike our other invariants,
a {\em small} value of the average path length is an evidence in favor of
{\em excessive} interconnection probability,
so we use the negated value
$$F_9(G) = -\mathtt{APL}(G)$$
in this work.
(If no path exists between $u$ and $v$, we use $s(u,v) = 2 \max s(u',v')$,
where the maximum is taken over all pairs of vertices that have an
existing path between them.)\footnote{
{\color{black}
In fact, the average path length (APL) is inappropriate for sparse
(highly disconnected) graphs. 
}}

\subsection{Temporal Normalization}
\label{sec:temp-norm}

\begin{figure}[!t]
  \centering
  \includegraphics[width=3.5in]{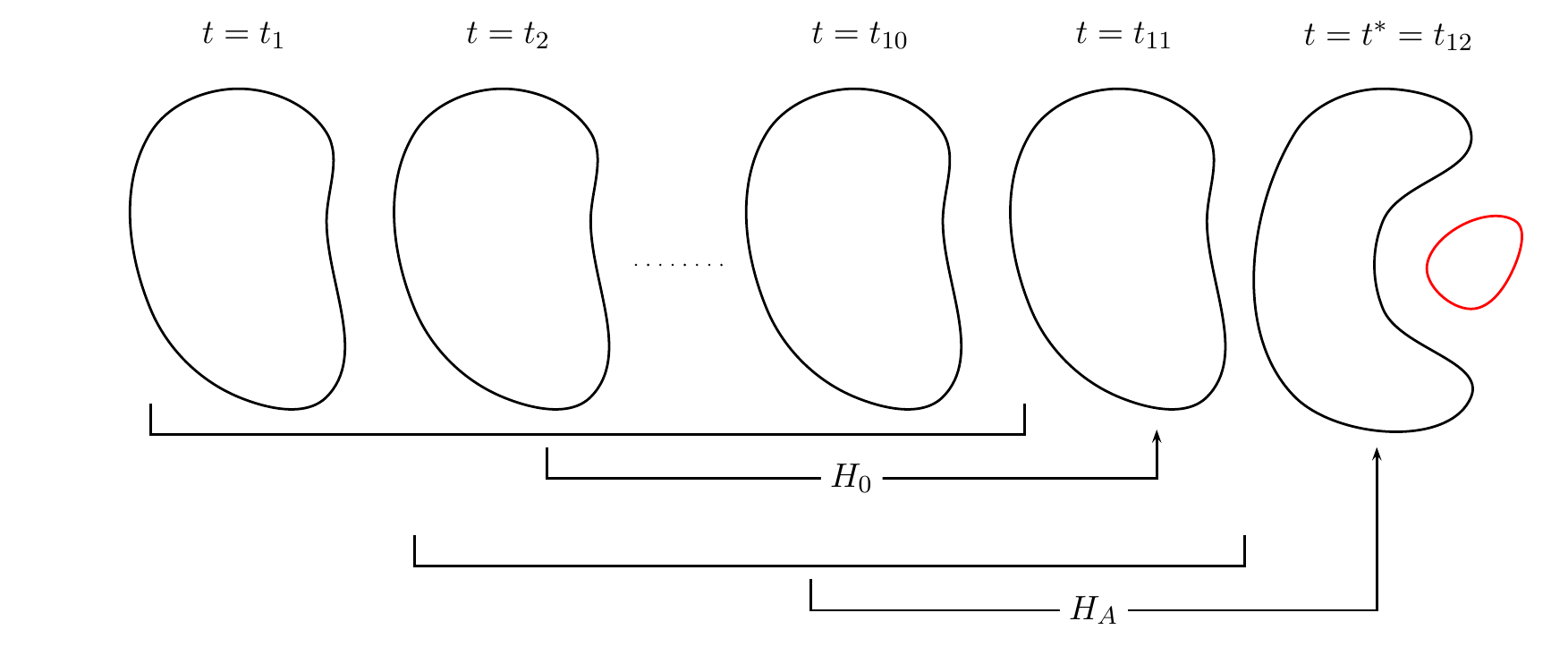}
  \caption{\label{fig:ke-ts-t}$H_0$ at $t=t^*-1$ and $H_A$ at
    $t=t^*$. {\color{black}The $H_0$ state compares previous many (10 in
      this case) null graphs to a
    null graph, $G(t=t_{11})$ and the $H_A$ state compares many null
    graphs to an alternative graph, $G(t=t^*=t_{12})$}.
  }
\end{figure}

The purpose of our inference is to detect a local (temporal) behavior 
change in 
the time series of graphs. In particular, we wish to consider as our
alternative hypothesis that a small (unspecified) collection of
vertices (the ``egg'') increases their within-group activity at some
time $t^*$ as compared to recent past 
while the majority of vertices (the ``kidney'')
continue with their normal behavior. The null hypothesis, then,
is a form of 
temporal homogeneity -- no probabilistic behavior changes in terms of
graph features. See Figure \ref{fig:ke-ts-t}. 

As mentioned in \cite{enron2005}, the raw features $F_{i}(G(t))$ are
standardized using a quantity computed from the recent past:
$$S_{i}(t) = \frac{F_{i}(G(t))-\widetilde{\mu}_{i,\ell}(t)}{\widetilde{\sigma}_{i,\ell}(t)},$$
where $\widetilde{\mu}_{i,\ell}(t)$ and $\widetilde{\sigma}_{i,\ell}(t)$
are the running mean and standard deviation estimates of
$F_{i}$ based on the most recent $\ell$ time steps; that is,
$$ \widetilde{\mu}_{i,\ell}(t) =
\frac{1}{\ell}\sum_{t'=t-\ell}^{t-1}F_{i}(G(t'))$$
and
$$\widetilde{\sigma}^{2}_{i,\ell}(t) = \frac{1}{\ell-1}\sum_{t'=t-\ell}^{t-1} (F_{i}(G(t'))-\widetilde{\mu}_{i,\ell}(t))^2.$$
 Then, a detection
at time $t$ is obtained when $S_{i}(t)$ is large.
(Note that for the localized statistics (maximum degree, maximum average degree, and the scan statistics)
we must first perform {\em vertex standardization}, as in \cite{enron2005} Section 6,
so that, for an {\em inhomogeneous} collection of stationary null vertex processes,
the most active vertices do not dominate these statistics.)

\subsection{Simulation}
\label{sec:synthetic-data}

Our general algorithm for
implementing the time series of random dot product graphs
is presented in
Algorithm \ref{alg:rdpg}.
The only difference among our three models in \cite{LP2009} occurs in line 3,
where the probability vectors for vertices are obtained;
the first approximation uses fixed ({\color{black}non-random or
  deterministic}) probability vectors $\pi_0$ and $\pi_A$ 
so that $\langle\piZbar,\piZbar\rangle$ $=$ $\langle\piZbar,\piAbar\rangle$ $=$ $p$ and $\langle\piAbar,\piAbar\rangle$ $=$ $q$
while
the second approximation and the exact models use random probability vectors
 \cite{LP2009}.

\begin{algorithm}[!t]
\caption{Time Series of Random Dot Product Graph}
\label{alg:rdpg}
\begin{algorithmic}[1]
\REQUIRE $n,\pi_0,\pi_A,t_{max}$
\FORALL {time $t$ such that $0 < t \leq t_{max}$}
   \STATE initialize the $n\times n$ adjacency matrix $A_t$ with zeros
   \STATE $vp \gets$ calculate probability vectors for all vertices
   using $(\pi_0,\pi_A)$
   \FORALL {vertex $u$ such that $1 \le u \le n$}
       \FORALL {vertex $v$ such that $1 \le v \le n$}
           \IF {$u>v$}
           \STATE $e \gets \langle {vp}_u,{vp}_v\rangle$
           \COMMENT{vector dot product}
           \STATE $A_t[u,v] \gets A_t[v,u] \gets Bernoulli(e)$
           \COMMENT{draw an edge}
           \ENDIF
       \ENDFOR
   \ENDFOR
   \STATE $A[t] \gets A_t$
\ENDFOR
\RETURN $A$, time series of graph
\end{algorithmic}
\end{algorithm}

\begin{figure}[!t]
  \centering
  \includegraphics[bb=0 0 500
  500,width=3.5in]{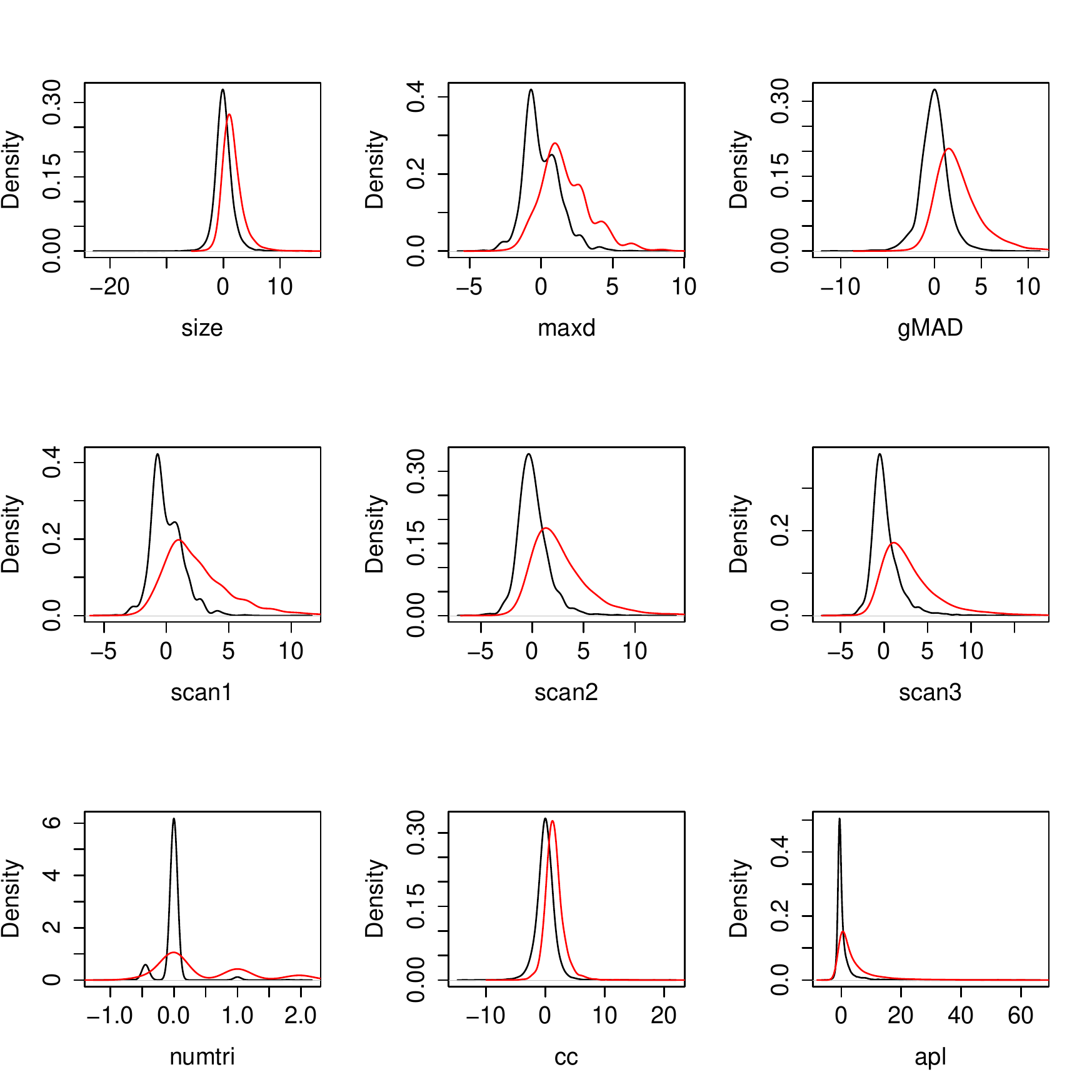}
  \caption{\label{fig:hist-ind}
    Density estimates for $M=10,000$ Monte Carlo replicates of $S_{i}(t)$
    in the first approximation model.
    $G(t)=ER(n=50,p=0.01)$ for $t=1,\cdots,t^*-1$
    and
    $G(t^*)=\kappa(n=50,p=0.01,m=6,q=0.3)$.
    For each invariant, black denotes $H_0: S_{i}(t^*-1)$ and red
    denotes $H_A: S_{i}(t^*)$.} 
\end{figure}

Density estimates of $S_{i}(t)$ for all nine features
are presented in Figure \ref{fig:hist-ind}
(using $\ell=5$).
Black denotes $H_0: S_{i}(t^*-1)$ and red denotes $H_A: S_{i}(t^*)$.
As we can see from this figure,
all features have mean zero and variance one (approximately) for $H_0$.
It is our goal to measure the performance of
each individual graph feature, and then compare these results with the
effectiveness of combining features, on our statistical inference task.

\begin{figure}[!t]
  \centering
  \includegraphics[bb=0 0 500 500,width=3.5in]{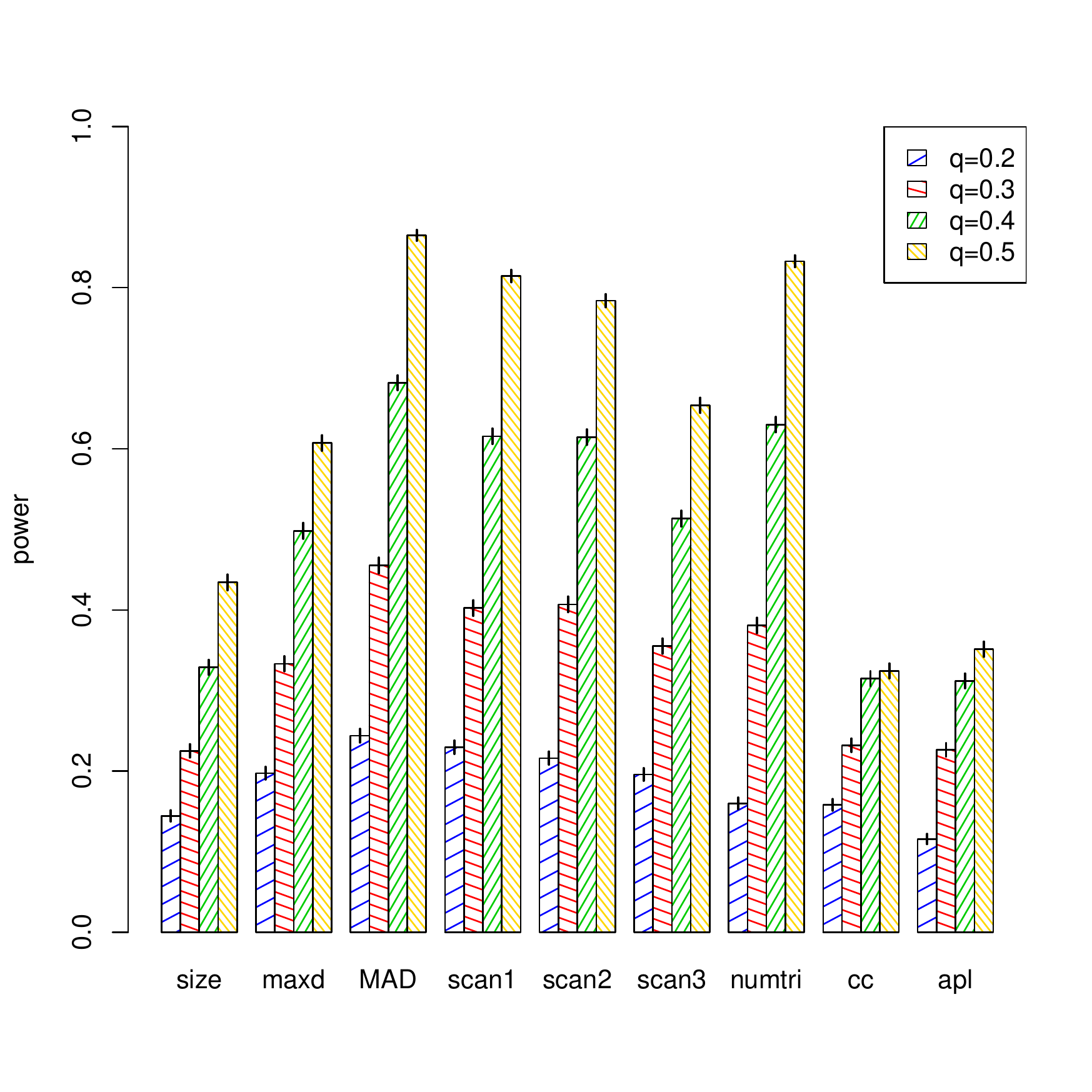}
  \caption{\label{fig:pwr-ind}
  Statistical power for our nine graph features
  in the first approximation model.
  $G(t)=ER(n=50,p=0.01)$ for $t=1,\cdots,t^*-1$
  and
  $G(t^*)=\kappa(n=50,p=0.01,m=6,q)$,
  for $q\in \{0.2,0.3,0.4,0.5\}$
  and allowable Type I error rate $\alpha=0.05$,
  based on $M=10,000$ Monte Carlo replicates.
  {\color{black} The error bars represent
        1.96 $\times$ 
      standard error for the sample means.}
  }
\end{figure}

Comparative power results for the individual features
are depicted in Figure \ref{fig:pwr-ind},
with a cumulative color bar  for each feature. 
For the most subtle case (when $q$ is small, in blue)
the power for each feature is relatively low,
while higher power is achieved as $q$ increases.
These results agree qualitatively with the results presented in \cite{pcp}.

\section{Fusion of Graph Features}
\label{sec:fusion}

We will consider two weighting methods for fusion of our graph features
introduced in  Section \ref{sec:graph-features}.
Our fusion test statistic is given by
$$S^w(t)=\sum_{i=1}^{d} w_{i}(t) S_{i}(t),$$
where $d$ is the number of graph features ($d=9$, for our investigations).

\subsection{Weighting}
\label{sec:weighting}

The naive equal weighting scheme is given by
$$w_{i}(t) = 1/d$$
for all $i$, and $t$.

Our \textit{adaptive} weighting scheme uses
$$w_{i}(t) = \frac{|S_{i}(t)-\mu_i(t)|}{\sigma_{i}(t)} \approx |S_{i}(t)|,$$
where $\mu_i(t)$ and $\sigma_i(t)$ are the mean and the standard deviation
of {\color{black} $S_i(t^*-1)$ over $M$ Monte Carlo replicates}.
(Due to our temporal normalization,
all features have mean zero and variance one (approximately)
when ``recent past'' consists of stationarity,
which is the assumption when testing for change at time $t$.)
A detailed algorithm of this approach is shown in
Algorithm \ref{alg:awf}.

\begin{algorithm}[!t]
\caption{Hypothesis Test using Adaptive Weighting Fusion}
\label{alg:awf}
\begin{algorithmic}[1]
\REQUIRE $S_i(t): M \times t_{max} \times d$ normalized feature
matrix, $t^*$
\STATE $S_i(t^*-1) \gets M \times d$ matrix for null at time $t^*-1$
from $S_i(t)$, and \\
 $S_i(t^*) \gets M \times d$ matrix for alternative at time $t^*$ from $S_i(t)$
\STATE $\mu_{0}(t) \gets 1 \times d$ mean vector of $S_i(t^*-1)$, and \\
$\sigma_{0}(t) \gets 1 \times d$ standard deviation vector of
$S_i(t^*-1)$ {\color{black} over $M$ Monte Carlo replicated}
\STATE $pwr \gets 0$
\FORALL {replicate $j$ such that $1 \leq j \leq M$}
   \STATE $x \gets S_i(t^*)[j,]$ \COMMENT{single replicate of $S_i(t^*)$}
   \STATE $w \gets |x - \mu_0(t)| / \sigma_0(t)$ \COMMENT{$1\times d$
     weight vector} 
   \STATE $S^w(t^*-1) \gets \sum_i^d w_iS_i(t^*-1)$ \COMMENT{$1 \times M$ fused
     null vector}
   \STATE $cv \gets$ quantile$(S^w(t^*-1),0.95)$ \COMMENT{critical value:
     95\% quantile}
   \STATE $S^w(t^*) \gets \sum_i^d w_ix_i$ \COMMENT{fused scalar of $x$}
   \IF {$S^w(t^*)>cv$}
   \STATE $pwr \gets pwr + 1$
   \ENDIF
\ENDFOR
\RETURN $pwr/M$, power of the test
\end{algorithmic}
\end{algorithm}

Notice that the adaptive weights are a function of the graph $G(t)$ being tested
(line 6 of the algorithm).
This implies that the features with larger deviations from the norm
get higher weights and contribute more to the inference.

\subsection{Examples}
\label{sec:examples}

\begin{figure*}[!t]
  \centering
  \subfigure[\small{Equal
    Weighting}]{\includegraphics[width=2.75in]{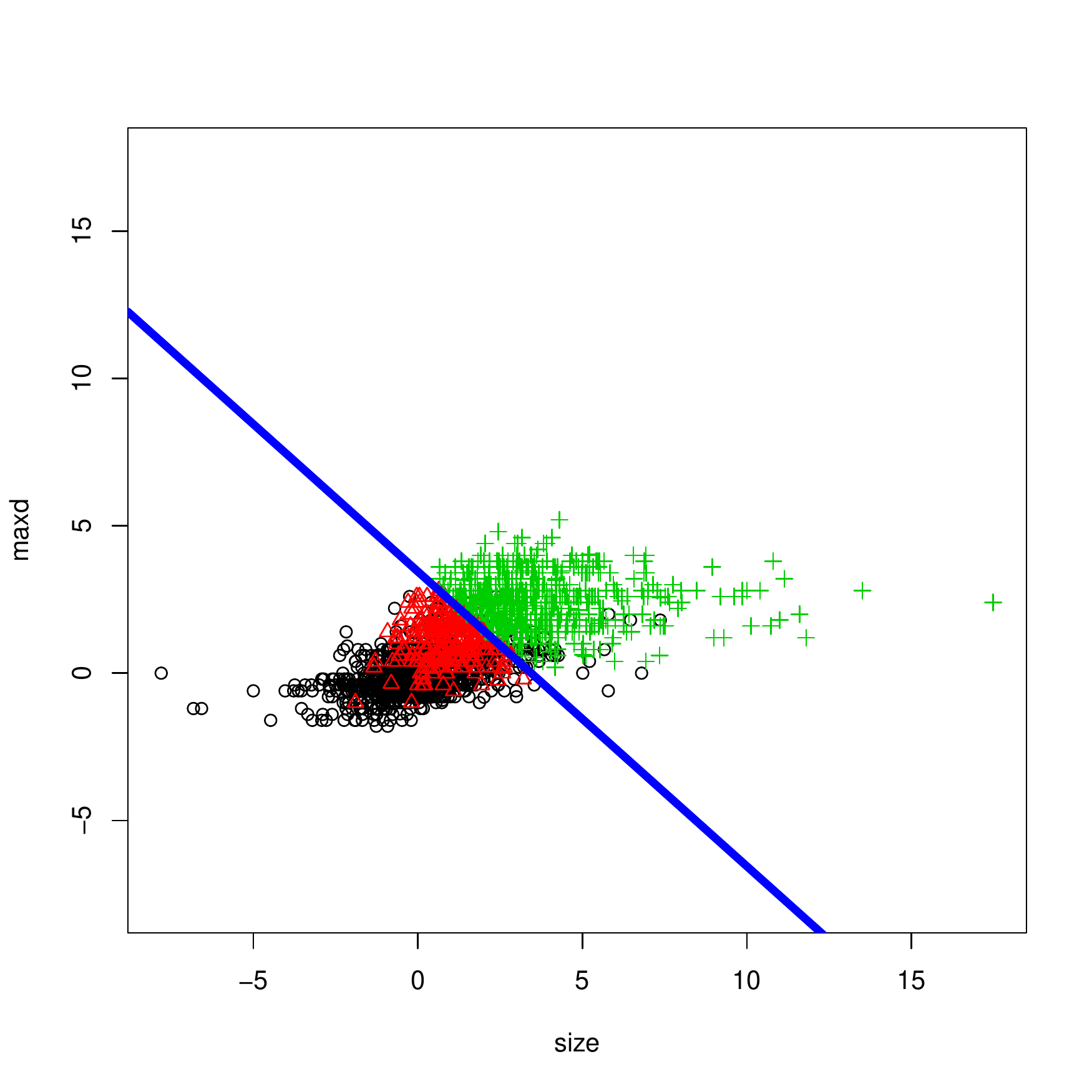}}
  \hfil 
  \subfigure[\small{Adaptive Weighting}]{\includegraphics[width=2.75in]{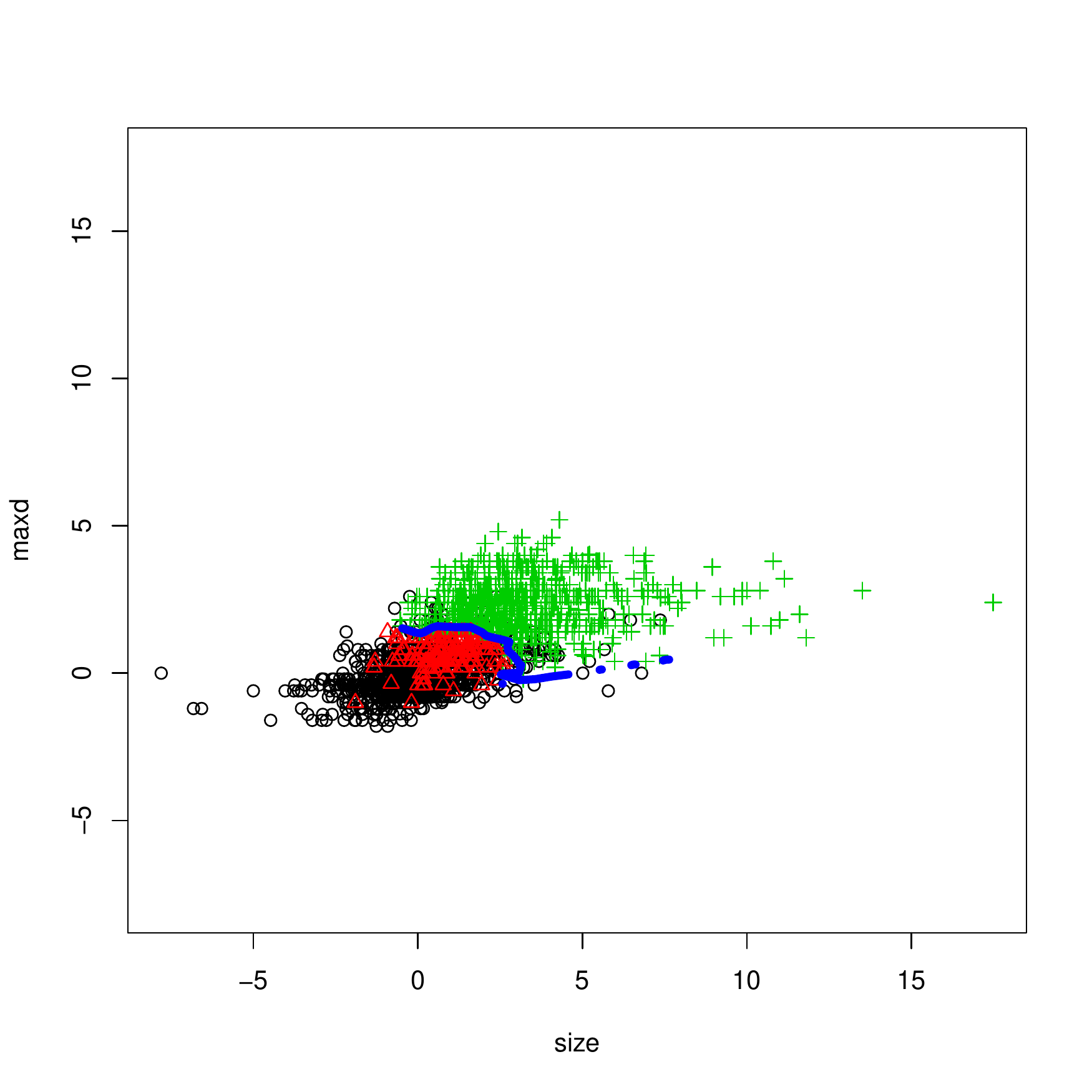}}
    \caption{\label{fig:scatter-fus1}Scatter plots for 
      size versus maximum degree for each fusion
      technique. Each point represents a Monte Carlo replicate. The black points
      (circles) are $S_i(t^*-1)$, and the color points 
      are $S_i(t^*)$; the points above the detection boundaries
      (critical values) are colored
      in green (``$+$'' symbols). {\color{black}The ratio of the number of green
      points over the total of green and red points represents the
      power of the test: power = 0.457 for the equal weighting and
      power = 0.564 for the adaptive weighting.} Blue lines represent 
      detection boundaries, which provide \textit{quantitative}
      rejection regions.} 
\end{figure*}

\begin{figure*}[!t]
  \centering
  \subfigure[\small{$q=0.2$}]{\includegraphics[width=2.75in]{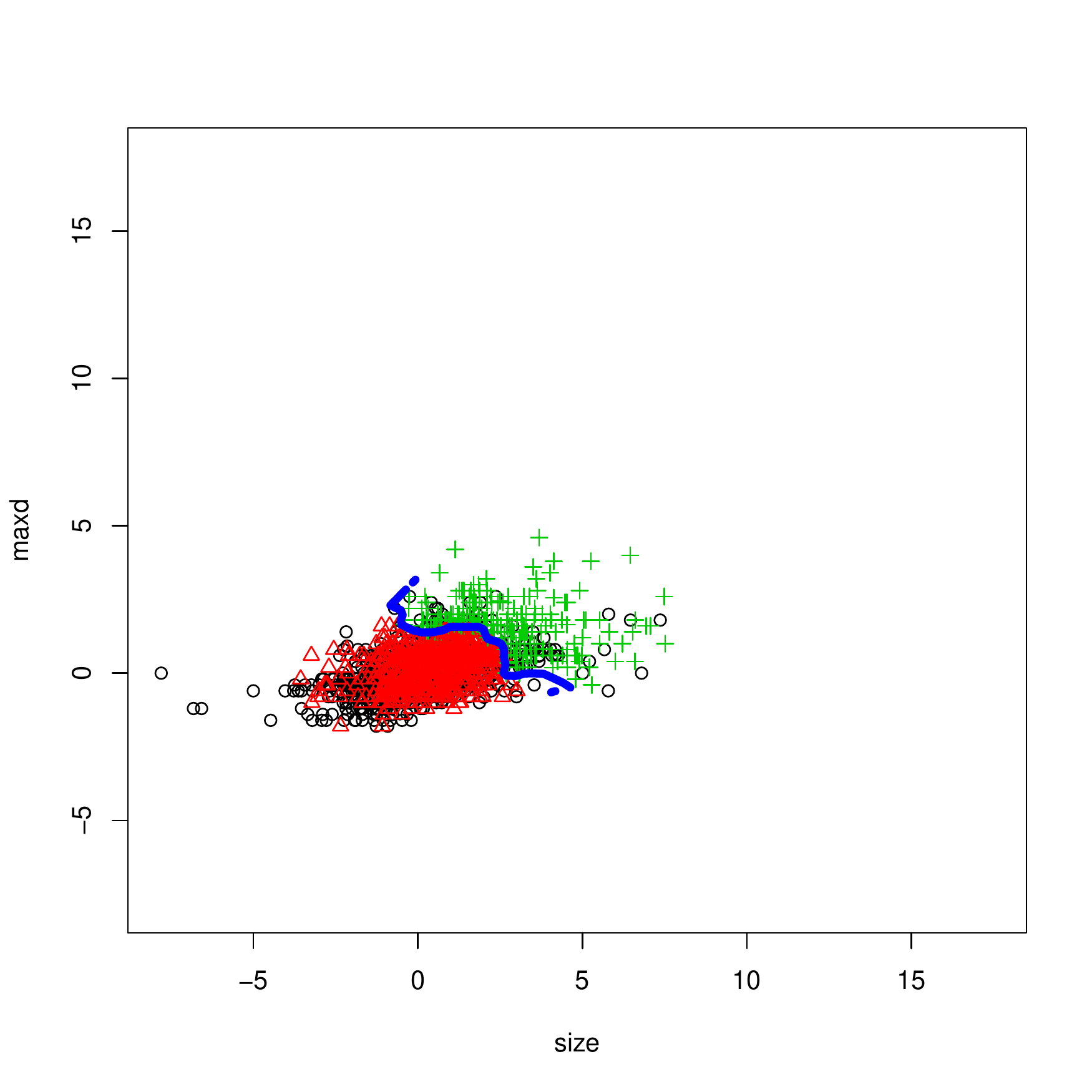}} \hfil
  \subfigure[\small{$q=0.3$}]{\includegraphics[width=2.75in]{scatter-fus4-q3.pdf}} 
  \subfigure[\small{$q=0.4$}]{\includegraphics[width=2.75in]{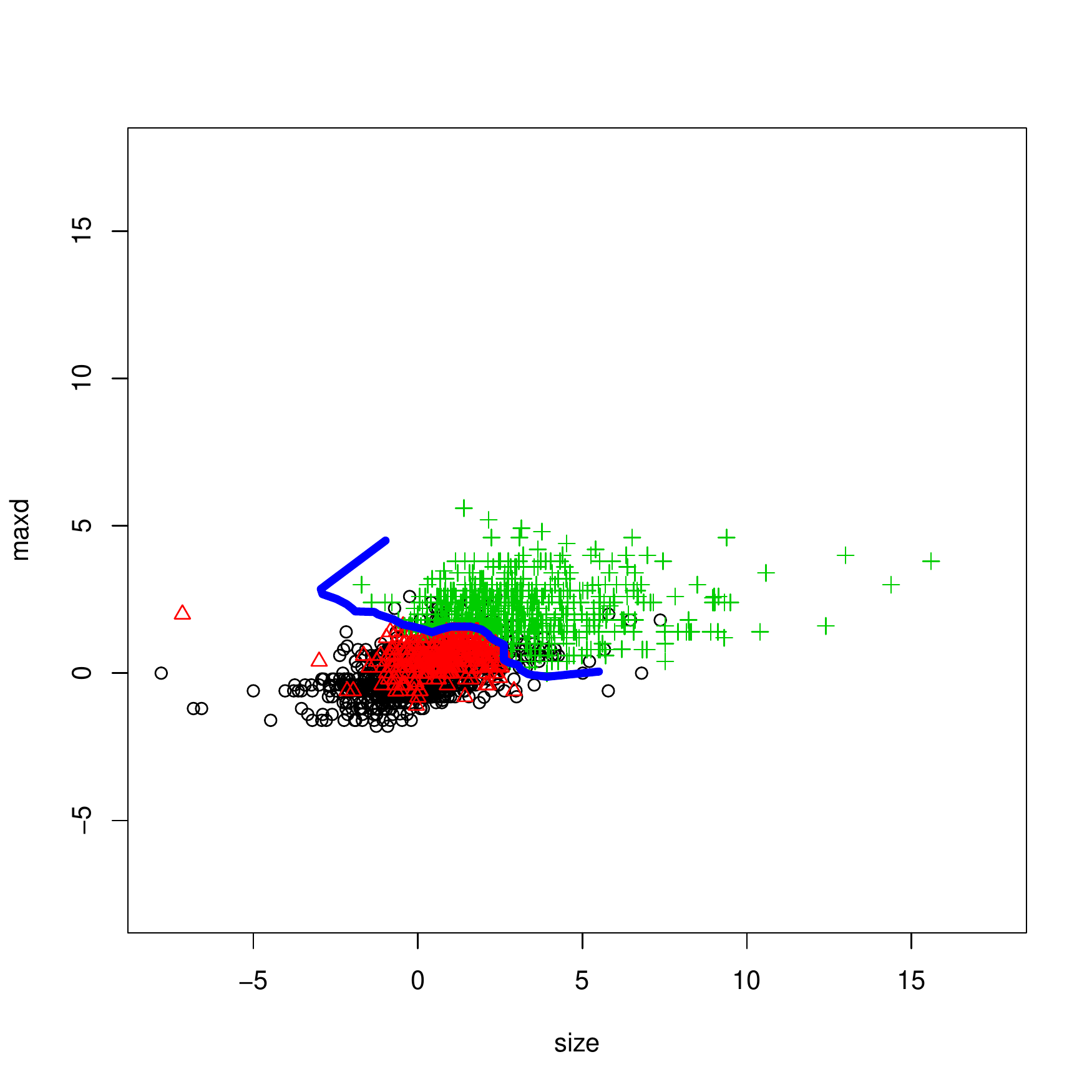}}
  \hfil 
  \subfigure[\small{$q=0.5$}]{\includegraphics[width=2.75in]{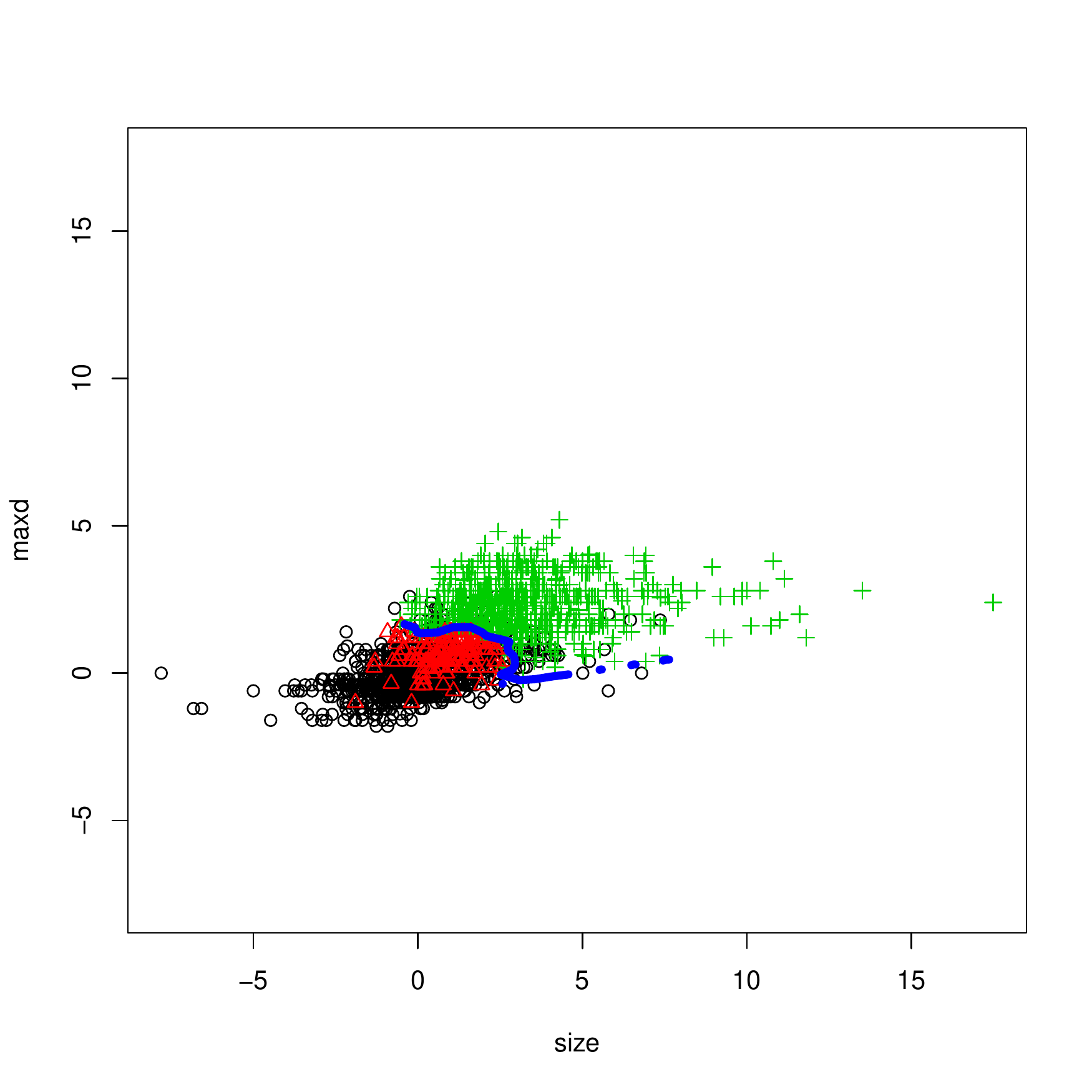}}
  \caption{\label{fig:scatter-fus2}Scatter plots for 
    size vs. maximum degree for adaptive
    weighting for
    $q=\{0.2,0.3,0.4,0.5\}$. 
    Each point represents a Monte Carlo replicate. The black points
    (circles) are $S_i(t^*-1)$, and the color points 
    are $S_i(t^*)$; the points above the detection boundaries
    (critical values) are colored
    in green (``$+$'' symbols). 
    {\color{black}The actual
    powers of the test are 0.332, 0.564, 0.775, and 0.917, respectively.}
    As $q$ increases, there are more green 
    points (``$+$'' symbols), which implies higher power. 
    Blue lines represent detection boundaries, which provide
    \textit{quantitative} rejection regions. 
  }
\end{figure*}

A graphical example is illustrated in Figures \ref{fig:scatter-fus1}
and \ref{fig:scatter-fus2}. In Figure \ref{fig:scatter-fus1}, each
point represents a Monte Carlo replicate of time series of graph in
two-dimensional 
Euclidean space using the first two features (size and maximum degree). 
The black points (circles) are 
$H_0:S_i(t^*-1)$, and the color points 
are $H_A:S_i(t^*)$; the points above the detection boundaries (critical
values in Algorithm \ref{alg:awf}, line 8) are colored
in green (``$+$'' symbols) and represent the power of the test. 
Notice that this boundary is linear for the equal weighting while
it is not for the adaptive weighting. The former is because the boundary is
calculated based on equal weighting for all $S_i(t^*-1)$ points; the
slope of the line is always -1 and the intercept can be calculated
with a given significance level of the test (\textit{i.e.,}
$ax+by>c, ~a=b=1/d, ~\therefore y>-x+dc, ~\text{where}~c=cv$). For the adaptive weighting case, meanwhile, the color
of the $S_i(t^*)$ points are determined by the distance from each point
to $\mu_0$, the mean vector of $S_i(t^*-1)$; the points whose fused
values are bigger than the 
critical value will get the green colors. 
This means that every $S_i(t^*)$ point gets a different
weight and therefore 
the detection boundary is not linear. Figure \ref{fig:scatter-fus2}
shows the adaptive weighting case  
for various values of $q$. As $q$ increases, there are more green
points, which implies higher power as expected.

\section{Fusion Experiments}
\label{sec:experiments}

\subsection{Simulations}
\label{sec:synthetic-data2}

\begin{figure}[!t]
  \centering
  \includegraphics[bb=0 0 500 500,width=3.5in]
    {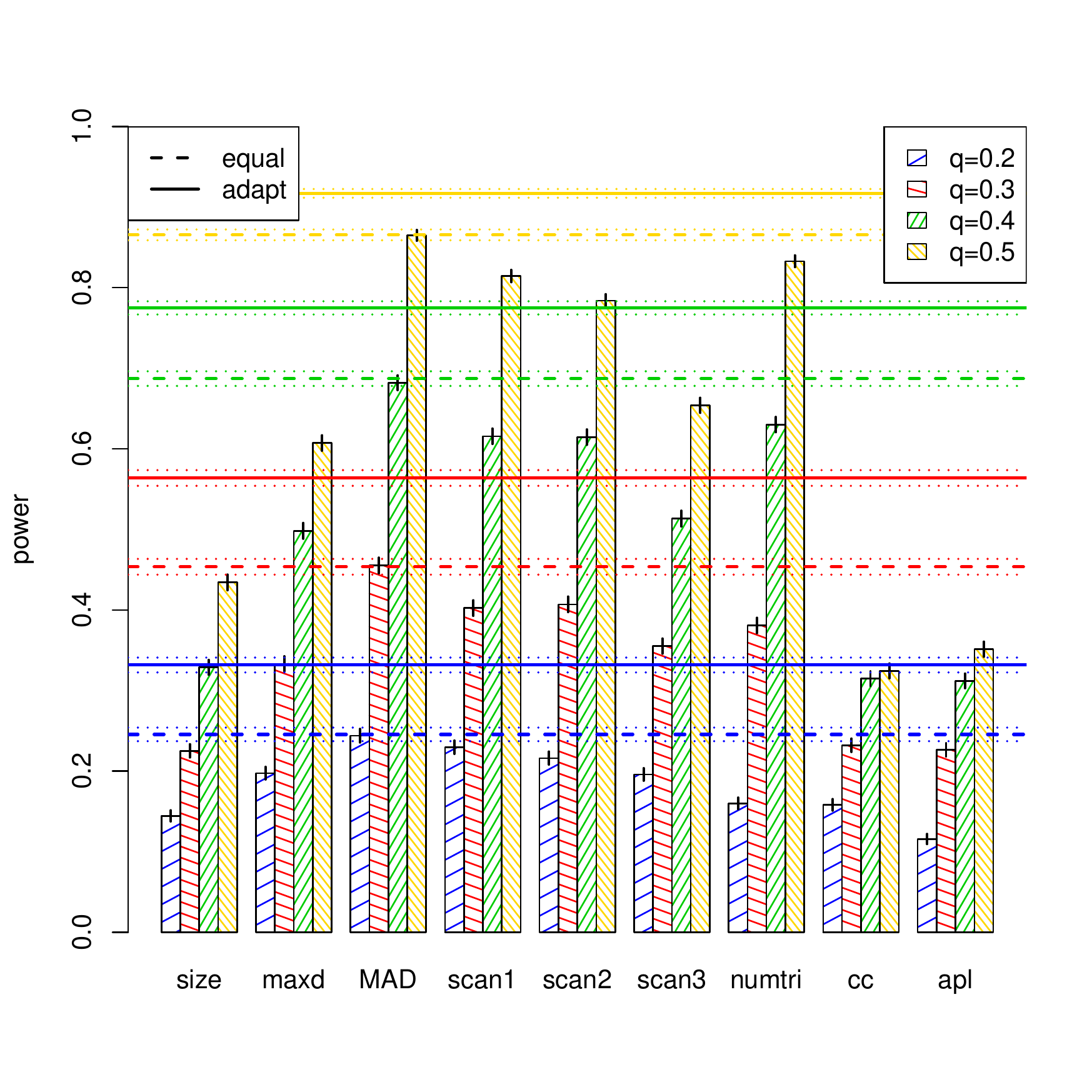}
  \caption{\label{fig:pwr-fus}
  Statistical power for our nine graph features
  and two fusion schemes
  in the first approximation model.
  $G(t)=ER(n=50,p=0.01)$ for $t=1,\cdots,t^*-1$
  and
  $G(t^*)=\kappa(n=50,p=0.01,m=6,q)$,
  for $q\in \{0.2,0.3,0.4,0.5\}$
  and allowable Type I error rate $\alpha=0.05$,
  based on $M=10,000$ Monte Carlo replicates.
  The horizontal lines indicate the power using fusion statistics
  $S^w(t)$ with $d'=9$. {\color{black} The error bars represent
        1.96 $\times$ 
      standard error for the sample means.}
  The superiority of adaptive weighting (solid lines) over equal weighting (dashed lines) is apparent.
}
\end{figure}

{\color{black} The simulation setup of this experiment is the same as the one in
Section \ref{sec:synthetic-data} except that fusion of graph features is
applied. }
The performance of
fusion with all nine features is depicted as 
horizontal lines in Figure \ref{fig:pwr-fus}. In all
cases, the fusion lines are above the corresponding individual bars,
and the adaptive weighting fusion lines are above the equal
weighting fusion lines.

\begin{figure}[!t]
  \centering
    \includegraphics[bb=0 0 500 500,width=3.5in]
    {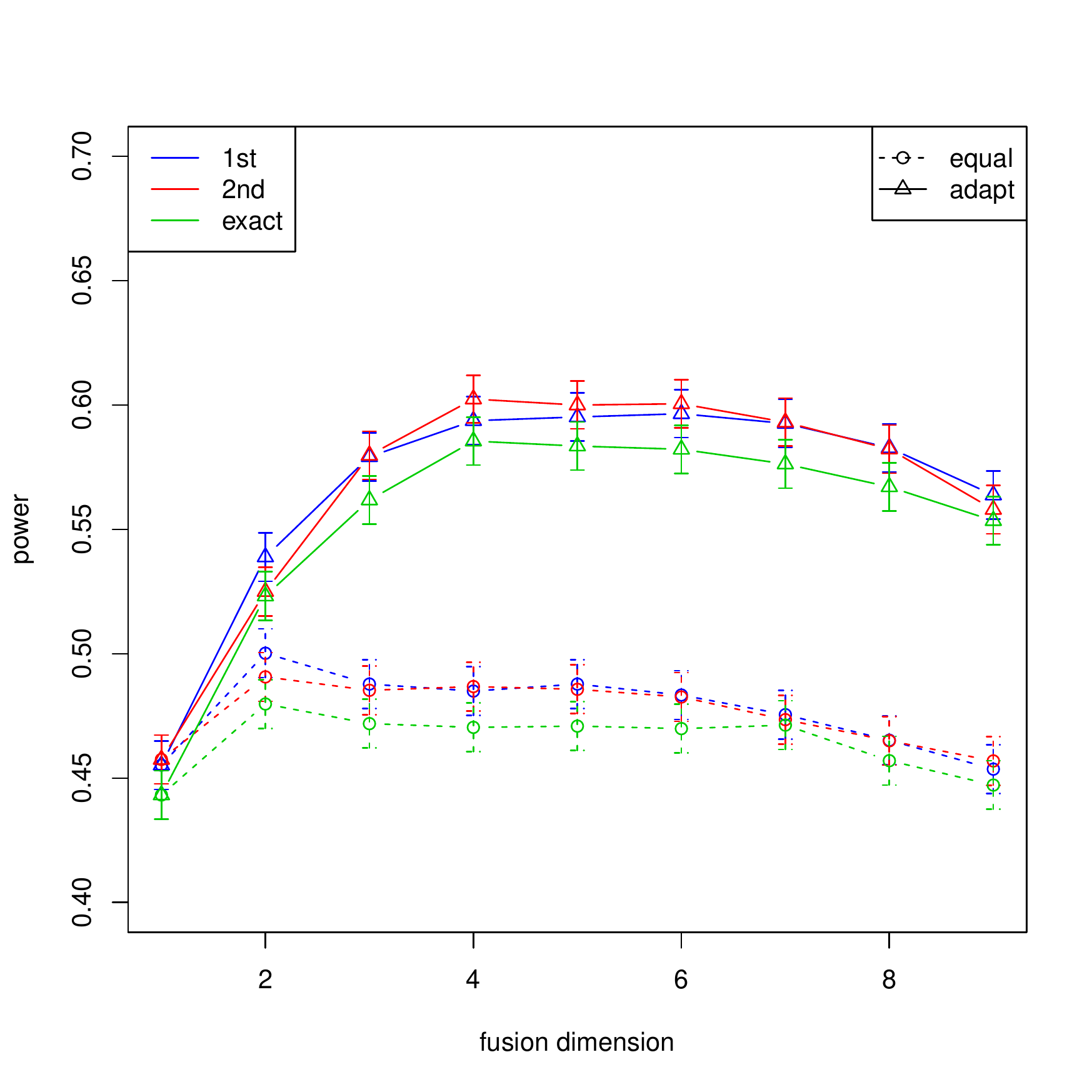}
    \caption{\label{fig:pwr-line}Statistical power plots for fusion
      statistics
      for the three models in \cite{LP2009}
       as a function of fusion dimension when $q=0.3$,
      $M=10,000$, and $\alpha=0.05$. {\color{black} The error bars represent
        1.96 $\times$ 
      standard error for the sample means. The fusion dimensions
      ($d'$) are chosen from the best possible combinations.}
      The difference in performance among the three models is minimal.
      The adaptive weighting scheme
      (with $\triangle$ symbol) is superior to equal weighting.
}
\end{figure}

Figure \ref{fig:pwr-line} depicts power
as a function of fusion dimension
for the different weighting schemes
for the three models in \cite{LP2009}.
Given a fusion dimension $d'$, all
${9\choose d'}$ possible combinations of features are considered for the
fusion and the best performance is plotted.
The difference in performance among
the three models in \cite{LP2009}
is minimal (``qualitatively similar''),
while the superiority of the adaptive weighting scheme (with $\triangle$
symbol) is apparent.
Table \ref{tab:pwr-d4} depicts the actual
weightings obtained via the adaptive 
weighting scheme for $d'=4$.
We see that, for the most part, the same features are selected for all
three models in \cite{LP2009}.

\begin{table}[!t]
  \caption{
    The estimated weightings obtained via the adaptive weighting scheme for $d'=4$ from Figure \ref{fig:pwr-line}.
    We see that, for the most part, the same features are selected for
    all three models in \cite{LP2009}.} 
  \label{tab:pwr-d4}
  \centering
  \begin{tabular}{lccccc}
    \toprule
    model      &  $\arg\max_i$  &  $w_1$ & $w_2$ & $w_3$ & $w_4$ \\
    \midrule
    1st approx &  (1,2,6,7)     &  2.66  & 0.86  & 1.30  & 0.10 \\
    2nd approx &  (1,2,6,7)     &  2.24  & 3.88  & 4.62  & 0.11 \\
    exact      &  (1,2,6,7)     &  1.25  & 5.14  & 6.01  & 13.9 \\
    \bottomrule
  \end{tabular}
\end{table}

\begin{figure}[!t]
  \centering
  \includegraphics[bb=0 0 500 500,width=3.5in]{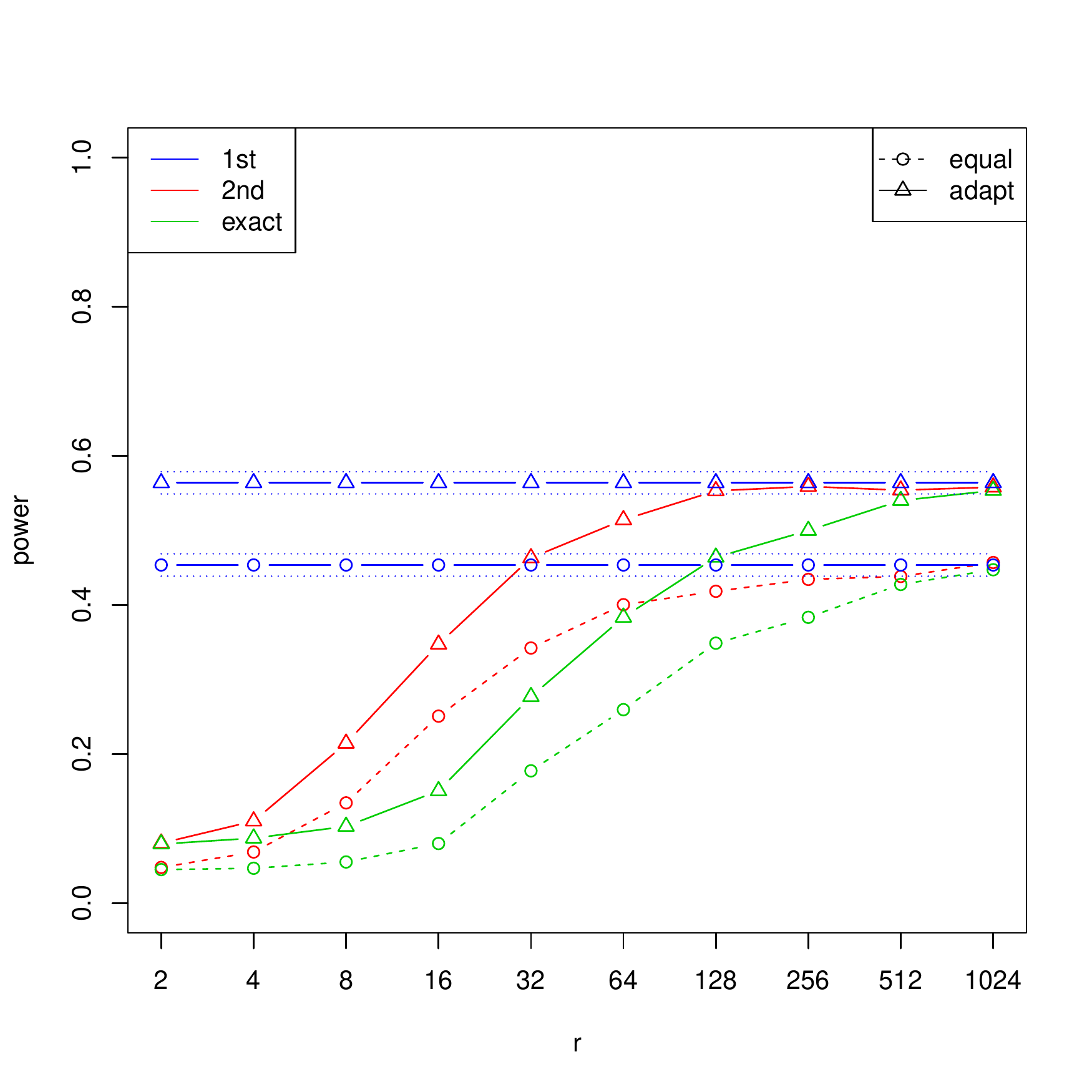}
  \caption{\label{fig:pwr-scale}
  Statistical power as a function of rate parameter $r$
  for models in \cite{LP2009} and both weighting schemes
  based on $M=10,000$ Monte Carlo replicates,
  with $d'=d=9$, $q=0.3$, and $\alpha=0.05$.
  The horizontal lines represent results for the first approximation ($r \to \infty$) $\pm$ three standard deviations
  for adaptive weighting (upper line, at power approximately 0.56)
  and equal weighting (lower line, at power approximately 0.45).
}
\end{figure}

In Figure \ref{fig:pwr-scale} we present a statistical power plot of fusion
using all nine features ($d'=d=9$)
with $q=0.3$ and $\alpha=0.05$
as a function of the rate parameter $r$ for the vertex processes\footnote{
{\color{black}The parameter $r$ controls the variability of the latent
  stochastic processes 
$\{X_v(t)\}$ for the vertices. In particular, a large value of $r$
corresponds to small 
variability in $\{X_v(t)\}$ (the second-order approximation), and as $r
\to \infty$ the processes $\{X_v(t)\}$ 
converge to the stationary probability vectors $\pi_0$ and $\pi_A$
(the first-order approximation).
See \cite{LP2009} for detail.
We have used $r=1024$ for all other results presented herein.}}.
These results demonstrate that
(1) adaptive weighting is superior to equal weighting,
(2) the second approximation is more faithful to the exact model than
is the first approximation, and
(3) both approximations are accurate for large $r$.

\subsection{Enron Email Data}
\label{sec:enron-email-data}

\begin{figure}[tbp]
    \centering
      \includegraphics[bb=0 0 500
      500,width=3.5in]{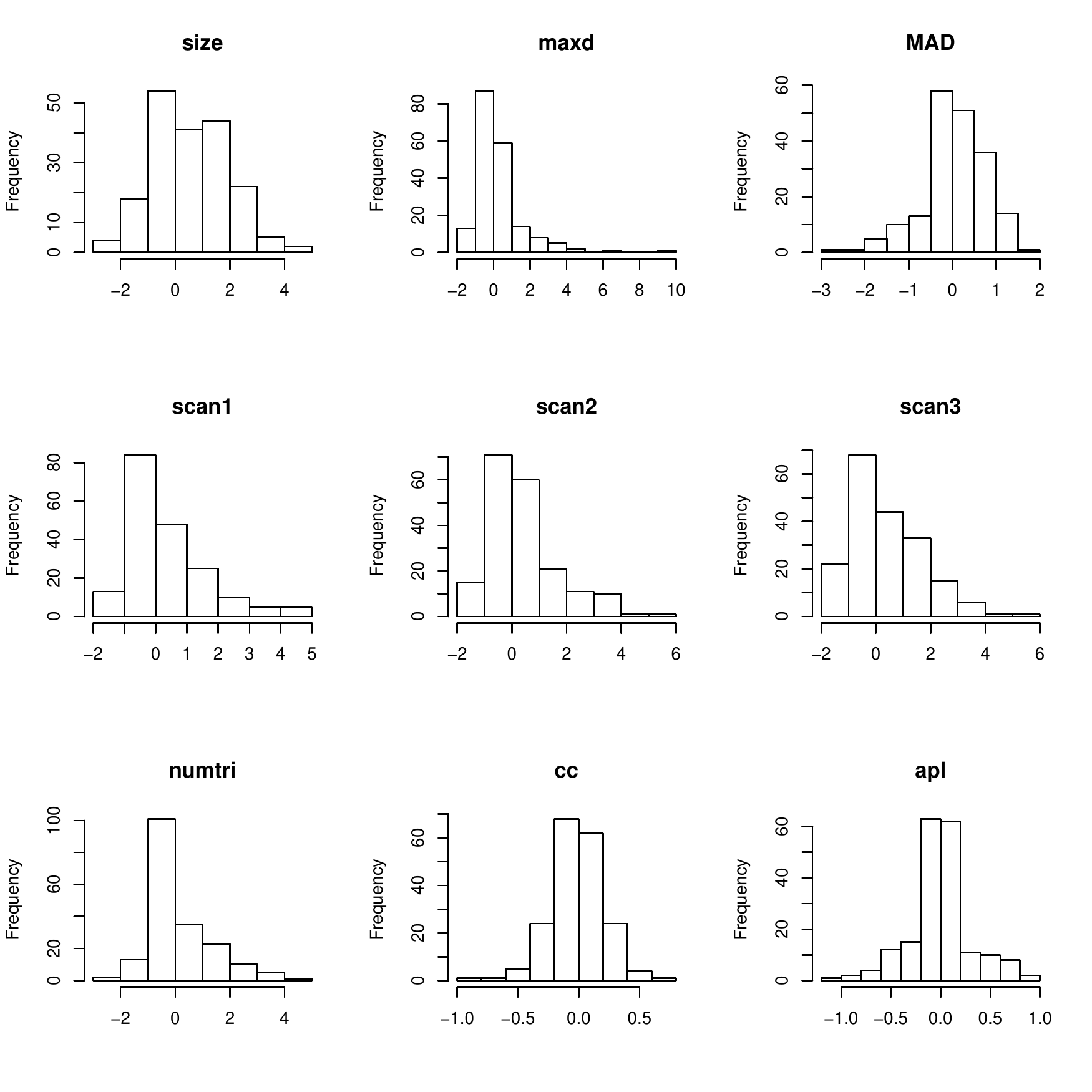}
    \caption{\label{fig:hist-enron}
    Enron email data
    histograms of $S_i(t)$
    for 189 weeks.
}
  \end{figure}

We use the Enron email data used in \cite{enron2005} for this
experiment. The nine features, $S_i(t)$ for $1 \le t \le
189$, are calculated for graphs derived from 
email messages among $n=184$ executives during one week periods.
Figure \ref{fig:hist-enron} depicts histograms of $S_i(t)$ for each $i$.

\begin{figure*}[!t]
    \centering
      \subfigure[\small{size vs. maxd}]{\includegraphics[width=2.75in]{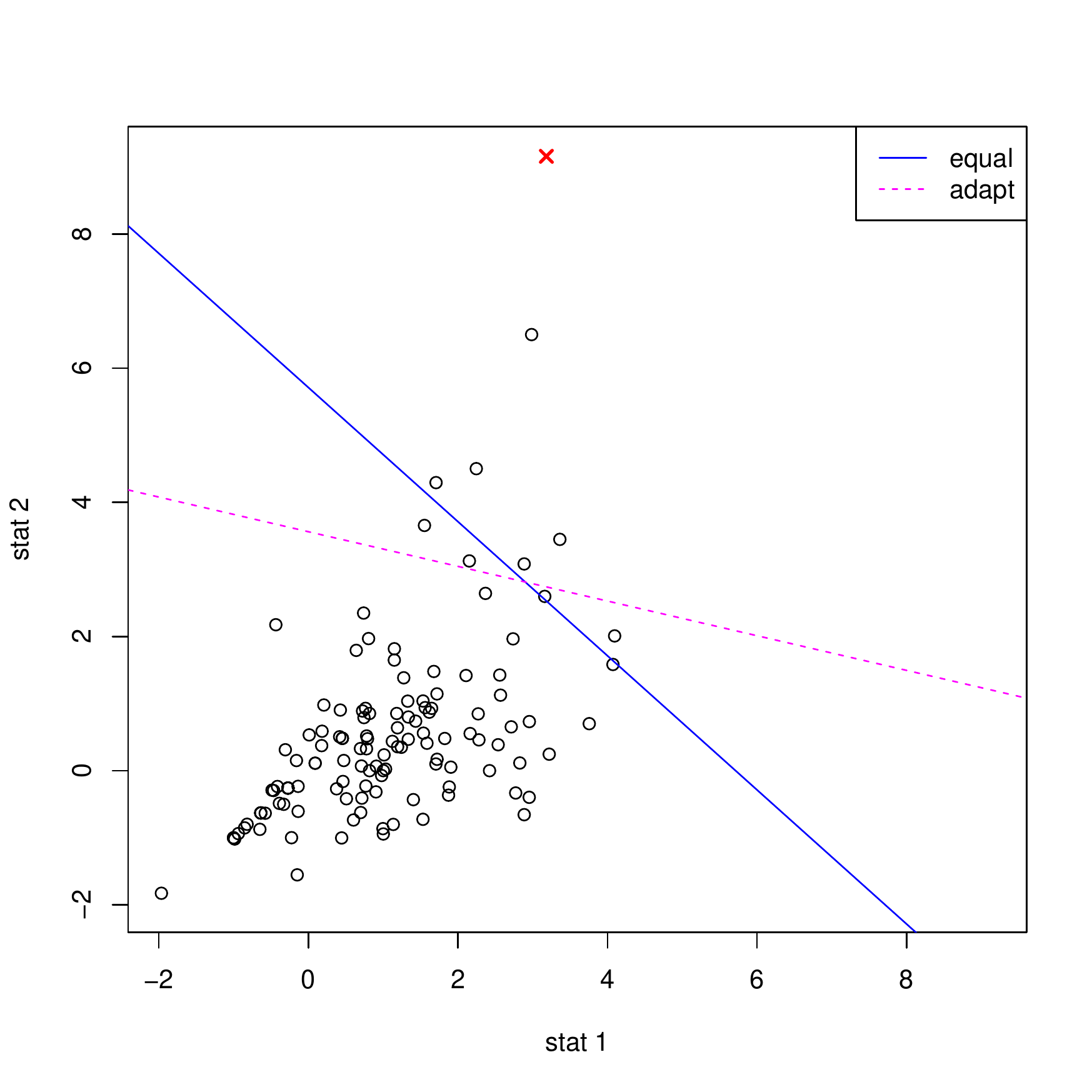}} \hfil
      \subfigure[\small{MAD vs. cc}]{\includegraphics[width=2.75in]{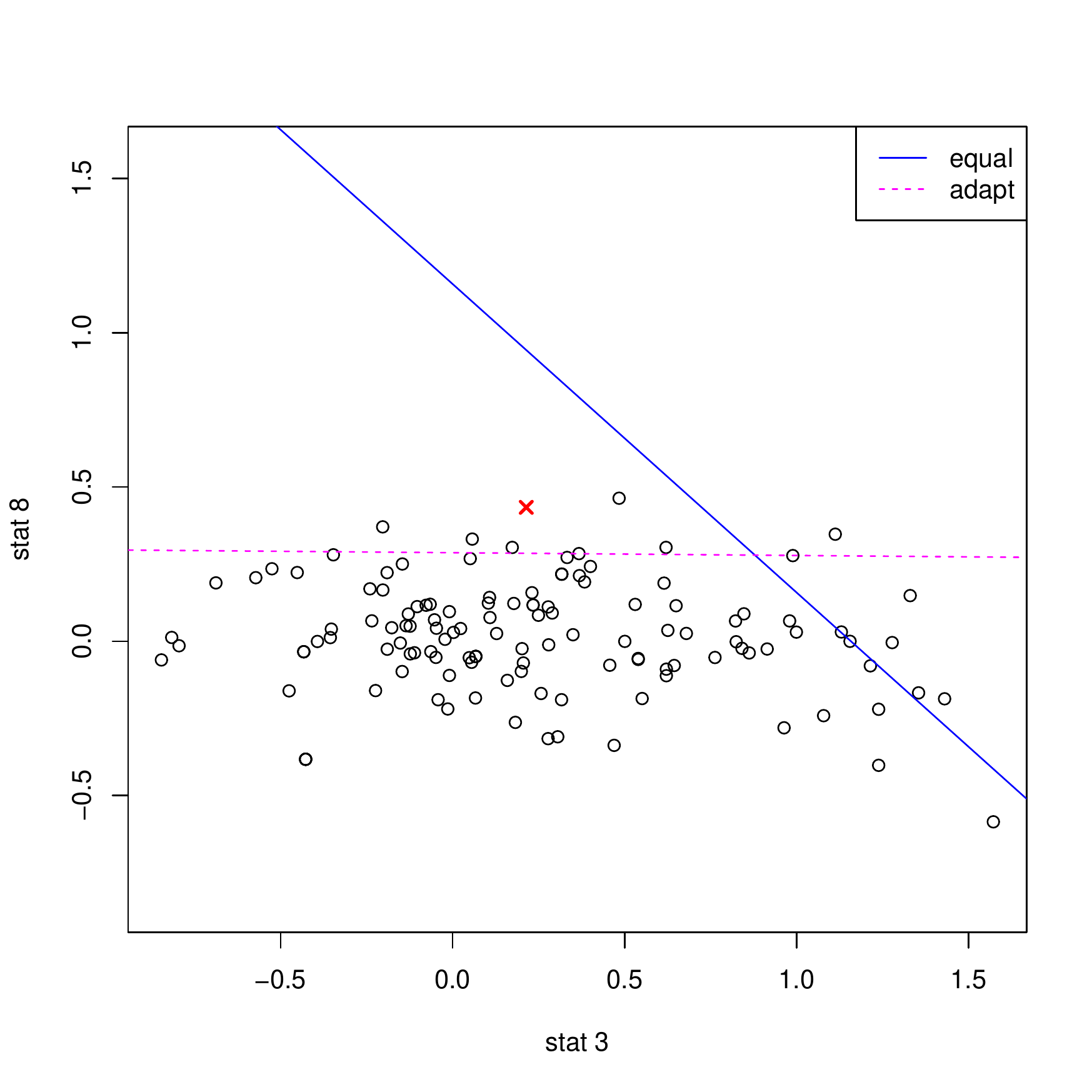}} 
      \subfigure[\small{scan1 vs. numtri}]{\includegraphics[width=2.75in]{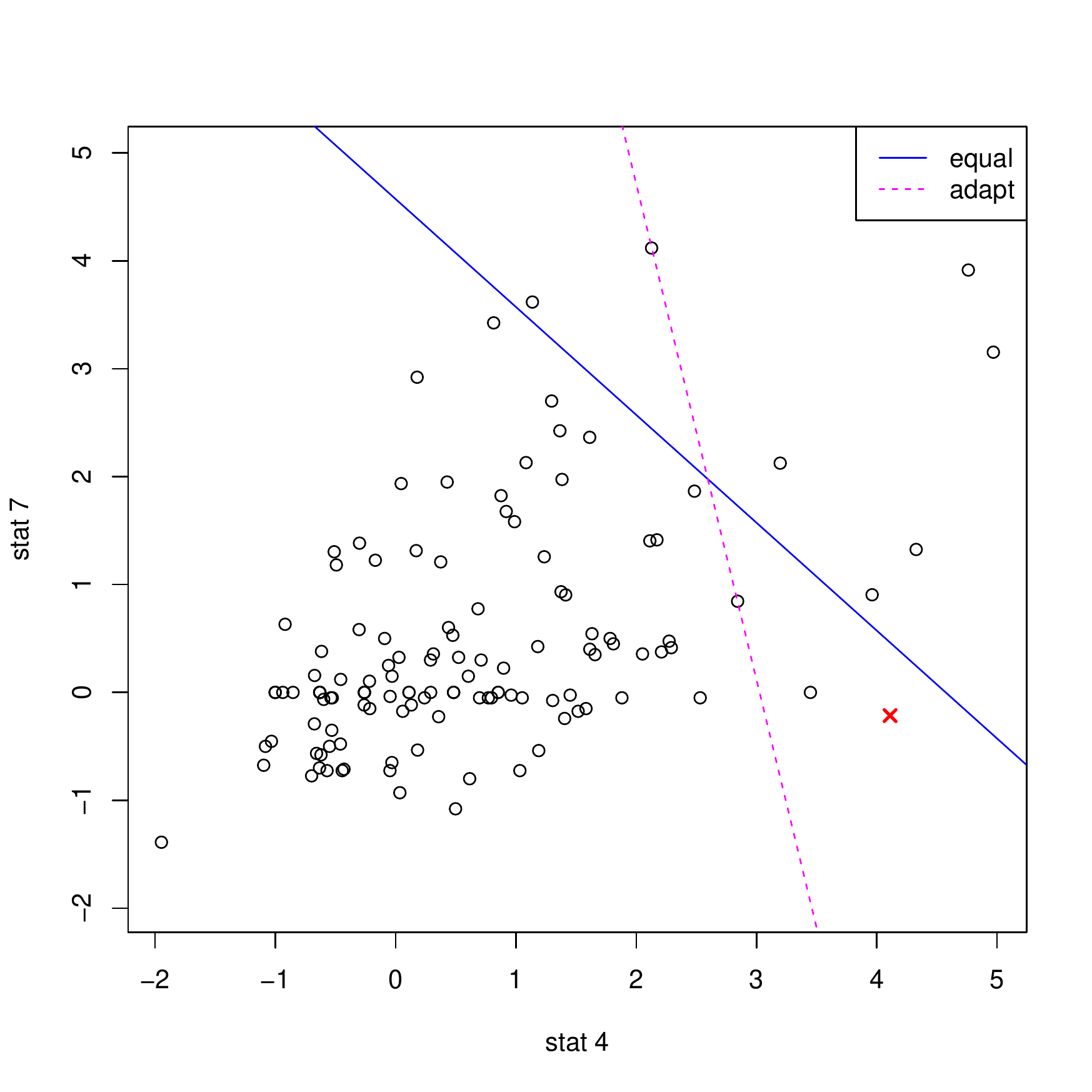}}
      \hfil 
      \subfigure[\small{scan3 vs numtri}]{\includegraphics[width=2.75in]{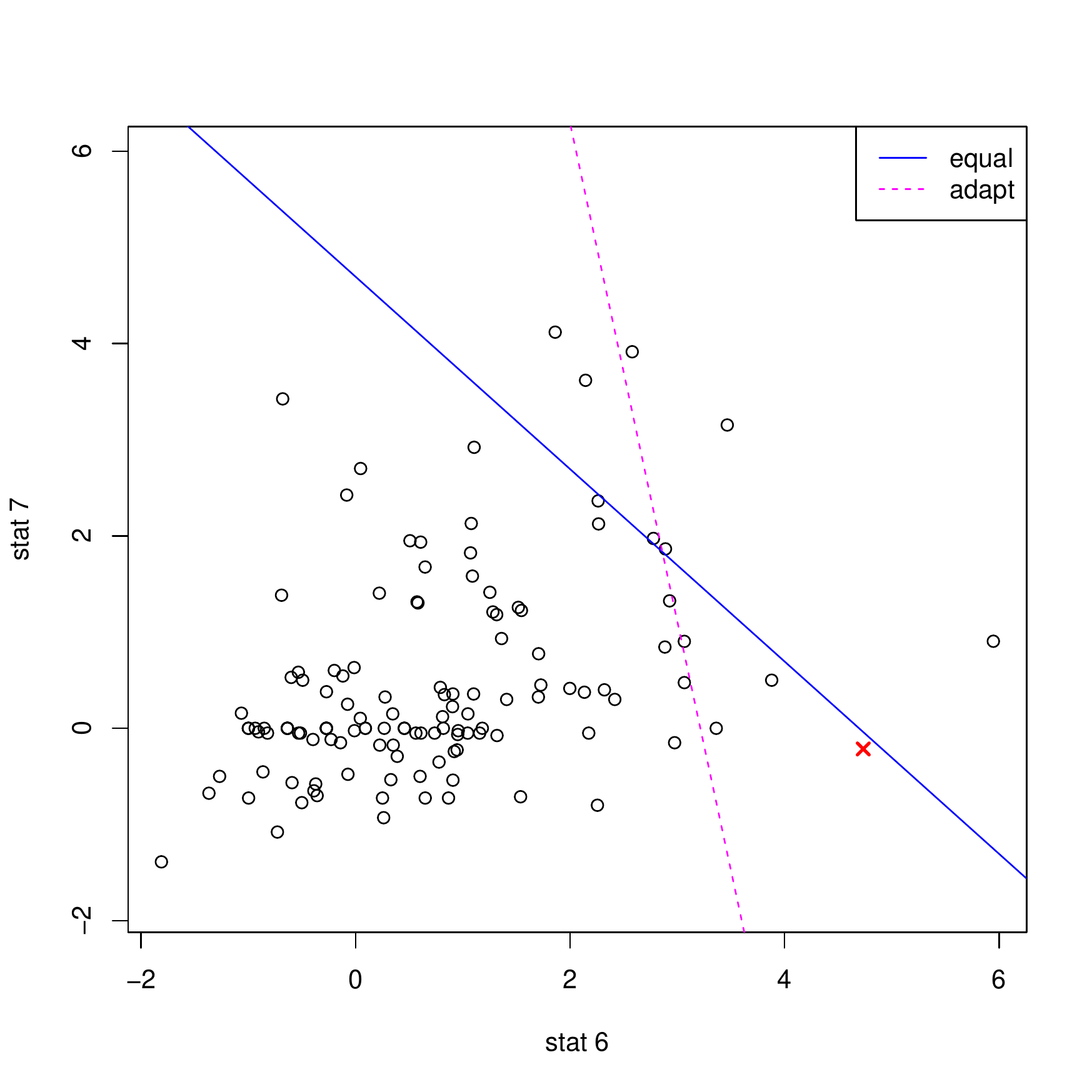}} 
    \caption{\label{fig:scatter-enron2}
      Enron email data scatter plots of $S_i(t)$ for
      $t=\{1,\ldots,132\}$ for various pairs of invariants.
      $S_i(t^*)$ is shown in red.
      The red point is above both critical lines in panel a,
      indicating that the combination of
      size and maximum degree allows detection based on $S_i(t^*)$
      for both weighting schemes.
      In panels b,c,d, it is apparent that only the adaptive weighting
      scheme detects the anomaly. Unlike Figure
      \ref{fig:scatter-fus2}, the detection boundaries for the
      adaptive weighting is linear, and it is because there is only
      one $S_i(t^*)$ graph.
    } 
 \end{figure*}

Our interest is the ``alias'' detection identified at week 132 in \cite{enron2005},
when an employee changes his/her email address.
Therefore, we choose $t^*=132$, the third week of May 2001.
Figure \ref{fig:scatter-enron2} depicts scatter plots of $S_i(t)$ for
$t=\{1,\ldots,132\}$ for various pairs of invariants, where
$S_i(t^*)$ is shown in red.
Unlike the simulation in Figure \ref{fig:scatter-fus2},
Monte Carlo replicates of graph are not available for real data;
therefore the 131 previous graphs (shown as black points in the figure)
are used to determine detection boundaries.
This investigation reveals that
the combination of
size and maximum degree allows detection based on $S_i(t^*)$
for both weighting schemes
(the red point is above both critical lines, in panel a),
while
only the adaptive weighting scheme detects the anomaly for the other three feature pairs depicted
(panels b,c,d).

{\color{black}The performance of equal and adaptive weighting fusion
  methods with all possible combinations of features at $t^*=132$ are
summarized in Table \ref{tab:enron}. For example, when the fusion dimension
$d'=2$, the possible number of combination of feature dimensions is 36, and
both equal and adaptive weighting methods can detect 24 cases, but
only adaptive weighting scheme can detect  5 additional cases. Note
that there is no case that only equal weighting scheme can detect
while adaptive weighting scheme cannot.}

\begin{table}[!t]
  \caption{The performance of equal and adaptive weighting fusion methods on
    Enron email graphs. For example, when the fusion dimension
    $d'=2$, the possible number of combination of feature dimensions is 36, and
    both equal and adaptive weighting methods can detect 24 cases, but
    only adaptive weighting can detect 5 additional cases.} 
  \label{tab:enron}
  \centering
  \begin{tabular}{crrrrrrrrr}
    \toprule
    $d'$ & 1 & 2 & 3 & 4 & 5 & 6 & 7 & 8 & 9 \\
    $\binom{9}{d'}$ & 9 & 36 & 84 & 126 & 126 & 84 & 36 & 9 & 1 \\
    \midrule
    both  & 6 & 24 & 65 & 106 & 116 & 81 & 36 & 9 & 1 \\
    equal & 0 & 0 & 0 & 0 & 0 & 0 & 0 & 0 & 0 \\
    \textbf{adapt} & 0 & \textbf{5} & \textbf{10} & \textbf{15} & \textbf{9} & \textbf{3} & 0 & 0 & 0 \\
    \bottomrule
  \end{tabular}
\end{table}

\section{Discussion}
\label{sec:discussion}
We have demonstrated,
via simulation results using a latent process model for time series of graphs
as well as illustrative experimental results for a time series of graphs derived from the Enron email data,
that an adaptive weighting methodology
for fusing information from graph features
provides superior inferential efficacy
for a certain class of anomaly detection problems.

One notable implication of this work is that
inferential performance in the mathematically tractable approximation
models in \cite{LP2009} 
does indeed provide guidance for methodological choices applicable to the exact (realistic but intractable) model.
Furthermore, to the extent possible, we may tentatively conclude that model investigations
have some bearing on real data applications.

An important extension of this work will be to
time series of \textit{weighted} and/or {\em attributed} graphs,
where message count and/or content is used to augment edges with
(categorical) ``topic'' attributes 
\cite{gpg,ppmcgg,tang:_attrib} where authors demonstrated that using
content and context together provides superior inferential capability
when compared to either alone for a number of inferential tasks.
{\color{black} Along with the fusion technique introduced in this paper,
  changes in communication {\em content}, in addition to 
excessive communication probability, can aid detection.}


%

\appendices
\ifCLASSOPTIONcompsoc
  \section*{Acknowledgments}
\else
  \section*{Acknowledgment}
\fi

\label{sec:acknowledgements}
This work was supported in part by the Johns Hopkins University Human
Language Technology Center of Excellence. The authors also would like to
thank the editors and the anonymous referees for their 
valuable comments and critiques that greatly improved this work. 

\ifCLASSOPTIONcaptionsoff
  \newpage
\fi



%



\bibliographystyle{IEEEtran}
\bibliography{IEEEabrv,parky}

\begin{thebibliography}{10}
\providecommand{\url}[1]{#1}
\csname url@samestyle\endcsname
\providecommand{\newblock}{\relax}
\providecommand{\bibinfo}[2]{#2}
\providecommand{\BIBentrySTDinterwordspacing}{\spaceskip=0pt\relax}
\providecommand{\BIBentryALTinterwordstretchfactor}{4}
\providecommand{\BIBentryALTinterwordspacing}{\spaceskip=\fontdimen2\font plus
\BIBentryALTinterwordstretchfactor\fontdimen3\font minus
  \fontdimen4\font\relax}
\providecommand{\BIBforeignlanguage}[2]{{%
\expandafter\ifx\csname l@#1\endcsname\relax
\typeout{** WARNING: IEEEtran.bst: No hyphenation pattern has been}%
\typeout{** loaded for the language `#1'. Using the pattern for}%
\typeout{** the default language instead.}%
\else
\language=\csname l@#1\endcsname
\fi
#2}}
\providecommand{\BIBdecl}{\relax}
\BIBdecl

\bibitem{LP2009}
N.~H. Lee and C.~E. Priebe, ``{A Latent Process Model for Time Series of
  Attributed Random Graphs},'' \emph{Statistical Inference for Stochastic
  Processes}, vol.~14, no.~3, pp. 231--253, 2011.

\bibitem{ST10}
\BIBentryALTinterwordspacing
E.~R. Scheinerman and K.~Tucker, ``{Modeling Graphs Using Dot Product
  Representations},'' \emph{Computational Statistics}, vol.~25, pp. 1--16,
  January 2010. [Online]. Available:
  \url{http://dx.doi.org/10.1007/s00180-009-0158-8}
\BIBentrySTDinterwordspacing

\bibitem{YS07}
\BIBentryALTinterwordspacing
S.~J. Young and E.~R. Scheinerman, ``{Random Dot Product Graph Models for
  Social Networks},'' \emph{Proceedings of the 5th International Conference on
  Algorithms and Models for the Web-Graph}, pp. 138--149, 2007. [Online].
  Available: \url{http://portal.acm.org/citation.cfm?id=1777879.1777890}
\BIBentrySTDinterwordspacing

\bibitem{BJR07}
B.~Bollob\'as, S.~Janson, and O.~Riordan, ``{The Phase Transition in
  Inhomogeneous Random Graphs},'' \emph{Random Structures and Algorithm},
  vol.~31, pp. 3--122, 2007.

\bibitem{HRH2002}
P.~Hoff, A.~E. Raftery, and M.~S. Handcock, ``{Latent Space Approaches to
  Social Network Analysis},'' \emph{Journal of the American Statistical
  Association}, vol.~97, pp. 1090--1098, 2002.

\bibitem{bollobas01}
B.~Bollob\'as, \emph{Random Graphs}, 2nd~ed.\hskip 1em plus 0.5em minus
  0.4em\relax Cambridge University Press, 2001.

\bibitem{enron2005}
C.~E. Priebe, J.~M. Conroy, D.~J. Marchette, and Y.~Park, ``{Scan Statistics on
  Enron Graphs},'' \emph{{Computational and Mathematical Organization Theory}},
  vol.~11, pp. 229--247, October 2005.

\bibitem{pcp}
H.~Pao, G.~A. Coppersmith, and C.~E. Priebe, ``{Statistical Inference on Random
  Graphs: Comparative Power Analyses via Monte Carlo},'' \emph{Journal of
  Computational and Graphical Statistics}, vol.~20, no.~2, pp. 395--416, 2011.

\bibitem{newsletter}
C.~E. Priebe, G.~A. Coppersmith, and A.~Rukhin, ``{You Say Graph Invariant, I
  Say Test Statistic},'' \emph{ASA Sections on Statistical Computing
  Statistical Graphics SCGN Newsletter}, vol.~21, no.~2, December 2010.

\bibitem{ide05}
T.~Ide and H.~Kashima, ``Eigenspace-based anomaly detection in computer
  systems,'' in \emph{Proceedings of the Tenth ACM SIGDD International
  Conference on Knowledge Discovery and Data mining}, 2005, pp. 440--449.

\bibitem{miller2011}
B.~A. Miller, M.~S. Beard, and N.~T. Bliss, ``Matched filtering for matched
  filtering for subgraph detection in dynamic networks,'' in \emph{Proc. IEEE
  Statistical Signal Processing Workshop (SSP)}, 2011.

\bibitem{borges2011}
N.~Borges, G.~A. Coppersmith, G.~G.~L. Meyer, and C.~E. Priebe, ``Anomaly
  detection for random graphs using distributions of vertex invariants,'' in
  \emph{2011 45th Annual Conference on Information Sciences and Systems
  (CISS)}, March 2011, pp. 1--6.

\bibitem{neil2011}
J.~Neil, C.~Storlie, C.~Hash, A.~Brugh, and M.~Fisk, ``Scan statistics for the
  online detection of locally anomalous subgraphs,'' \emph{Technometrics (in
  review)}, 2012.

\bibitem{horn2011}
C.~Horn and R.~Willett, ``Online anomaly detection with expert system feedback
  in social networks,'' in \emph{2011 IEEE International Conference on
  Acoustics, Speech and Signal Processing (ICASSP)}, May 2011, pp. 1936--1939.

\bibitem{sharpnack}
J.~Sharpnack, A.~Rinaldo, and A.~Singh, ``Changepoint detection over graphs
  with the spectral scan statistic,'' \emph{arXiv/1206.0773}, 2012.

\bibitem{valko}
M.~Valko, ``Adaptive graph-based algorithms for conditional anomaly detection
  and semi-supervised learning,'' Ph.D. dissertation, University of Pittsburgh,
  2011.

\bibitem{fgt97}
D.~Ullman and E.~R. Scheinerman, \emph{Fractional Graph Theory}.\hskip 1em plus
  0.5em minus 0.4em\relax Wiley, 1997.

\bibitem{gpg}
J.~Grothendieck, C.~E. Priebe, and A.~L. Gorin, ``{Statistical Inference on
  Attributed Random Graphs: Fusion of Graph Features and Content},''
  \emph{Computational Statistics and Data Analysis}, vol.~54, pp. 1777--1790,
  2010.

\bibitem{ppmcgg}
C.~E. Priebe, Y.~Park, D.~J. Marchette, J.~M. Conroy, J.~Grothendieck, and
  A.~Gorin, ``{Statistical Inference on Attributed Random Graphs: Fusion of
  Graph Features and Content: An Experiment on Time Series of Enron Graphs},''
  \emph{Computational Statistics and Data Analysis}, vol.~54, pp. 1766--1776,
  2010.

\bibitem{tang:_attrib}
M.~Tang, Y.~Park, N.~H. Lee, and C.~E. Priebe, ``Attribute fusion in a latent
  process model for time-series of graphs,'' 2012, submitted.

\end{thebibliography}





\end{document}